%% file: arXiv_SocialFraudNetworkSimulationEngine_CampoBavoDC.tex
\documentclass[11pt]{article}
\pdfmapfile{+bickham.map}
\pdfoutput=1
\parskip \medskipamount

\usepackage{natbib}
\usepackage{har2nat}
\setcitestyle{authoryear,open={(},close={)}}

\renewcommand*\cite[1]{\citet{#1}}

\usepackage[a4paper, left=1in, right=1in, top=1.8cm, bottom=1.8cm, includehead, includefoot]{geometry} 	
\usepackage{amsfonts}
\usepackage{array}
\usepackage{natbib}											
\usepackage{graphicx} 									
\usepackage[small,bf,hang]{caption}			
\usepackage[labelformat=simple]{subcaption}									

	\usepackage{wrapfig}										
	\usepackage{multirow}
	\usepackage[table]{xcolor}										
	\definecolor{lightblue}{rgb}{0.13, 0.67, 0.8}
	\definecolor{tableblue}{rgb}{250, 250, 255}
	\definecolor{kul-blue}{rgb}{29,141,176}	
	
	\usepackage[latin1]{inputenc}						
	\usepackage{amsmath}										
	\allowdisplaybreaks[1]									
	\usepackage{amssymb}										
	\usepackage{amsthm}	
	\usepackage{leftidx}										
	\usepackage{booktabs}										
	\usepackage{url}                        
	\usepackage{eurosym}                    
	\usepackage{textcomp}                   
	\usepackage{fancyhdr}                   
	\usepackage{listings} 									
	\usepackage{pdfpages}										
	\usepackage{floatrow}
	\usepackage[title,titletoc,toc]{appendix}				
		\usepackage{enumerate}									
		\usepackage[multiple,hang,flushmargin]{footmisc}																		
		\usepackage{enumitem}										
		\usepackage{mathtools}										
		\usepackage{IEEEtrantools}							
		\usepackage{authblk}											
		\usepackage{tikz}
		\usetikzlibrary{snakes}
		\usetikzlibrary{shapes,arrows,decorations.pathmorphing,backgrounds,positioning,fit,petri, matrix, arrows.meta}
		\usepackage[english]{babel}
		\usepackage{grffile}
		\usepackage{amsfonts}
		\usepackage{amssymb}
		\usepackage{relsize}
		\usepackage{rotating}
		\usepackage{siunitx}
		\usepackage{color}	
		\usepackage{etoolbox}
		\usepackage{todonotes}
		\usepackage{longtable}
		\usepackage{spbmark}
		\usepackage{pdflscape}
		\usepackage[plainpages=false,hyperfootnotes=false]{hyperref}
		\hypersetup{
			colorlinks   = true, 
			urlcolor     = blue, 
			linkcolor    = red, 
			citecolor   = blue 
		}
		\usepackage{kbordermatrix}
		\usepackage[capitalize]{cleveref}
		
		\usepackage{bookmark}
		\usepackage{afterpage}
		
		\usepackage[ruled,vlined,linesnumbered]{algorithm2e}
		\SetAlCapFnt{\small} 
		\SetKw{KwTo}{in}
		
		\SetKwInput{kwModel}{Model}
		\SetKwInput{kwInit}{Initialization}
		\newcommand{\nextnr}{\stepcounter{AlgoLine}\ShowLn}
		
		\usetikzlibrary{tikzmark,fit}
		\usetikzlibrary{decorations.pathreplacing,calc}

		\makeatletter
		
		\pretocmd{\NAT@citex}{%
			\let\NAT@hyper@\NAT@hyper@citex
			\def\NAT@postnote{#2}%
			\setcounter{NAT@total@cites}{0}%
			\setcounter{NAT@count@cites}{0}%
			\forcsvlist{\stepcounter{NAT@total@cites}\@gobble}{#3}}{}{}
		\newcounter{NAT@total@cites}
		\newcounter{NAT@count@cites}
		\def\NAT@postnote{}
		
		\def\NAT@hyper@citex#1{%
			\stepcounter{NAT@count@cites}%
			\hyper@natlinkstart{\@citeb\@extra@b@citeb}#1%
			\ifnumequal{\value{NAT@count@cites}}{\value{NAT@total@cites}}
			{\ifNAT@swa\else\if*\NAT@postnote*\else%
				\NAT@cmt\NAT@postnote\global\def\NAT@postnote{}\fi\fi}{}%
			\ifNAT@swa\else\if\relax\NAT@date\relax
			\else\NAT@@close\global\let\NAT@nm\@empty\fi\fi
			\hyper@natlinkend}
		\renewcommand\hyper@natlinkbreak[2]{#1}
		
		\patchcmd{\NAT@citex}
		{\ifNAT@swa\else\if*#2*\else\NAT@cmt#2\fi
			\if\relax\NAT@date\relax\else\NAT@@close\fi\fi}{}{}{}
		\patchcmd{\NAT@citex}
		{\if\relax\NAT@date\relax\NAT@def@citea\else\NAT@def@citea@close\fi}
		{\if\relax\NAT@date\relax\NAT@def@citea\else\NAT@def@citea@space\fi}{}{}
		
		\makeatother
		
		\definecolor{darkgreen}{RGB}{40,150,40}
		
		\newcommand{\bs}{\boldsymbol}
		\newcommand{\mc}{\mathcal}

		\renewcommand{\leq}{\leqslant}					
		\renewcommand{\geq}{\geqslant}					
		\renewcommand{\epsilon}{\varepsilon}

		\theoremstyle{definition}

		\usepackage[scr=bickham, scrscaled=0.95, cal = boondox, calscaled=1.1]{mathalpha}
		\makeatletter
		\newcommand{\raisemath}[1]{\mathpalette{\raisem@th{#1}}}
		\newcommand{\raisem@th}[3]{\raisebox{#1}{$#2#3$}}
		\makeatother

		\newcommand{\ignore}[1]{}

		\newcolumntype{K}{>{\centering\arraybackslash}m{1.5cm}}
		\newcolumntype{M}{>{\centering\arraybackslash}m{3.5cm}}
		\newcolumntype{B}{>{\raggedright\arraybackslash}m{5cm}}
		\newcolumntype{W}{>{\raggedright\arraybackslash}p{12cm}}
		\newcolumntype{X}{>{\raggedright\arraybackslash}p{8cm}}
		\newcolumntype{N}{>{\raggedright\arraybackslash}p{9.0cm}}
		\newcolumntype{C}[1]{>{\centering\let\newline\\\arraybackslash\hspace{0pt}}m{#1}}
		\newcolumntype{D}{>{\raggedright\arraybackslash}m{1cm}}
		\newcolumntype{E}{>{\raggedright\arraybackslash}m{1.5cm}}
		\newcolumntype{H}{>{\setbox0=\hbox\bgroup}c<{\egroup}@{}}
		\newcolumntype{I}{>{\centering\arraybackslash}m{2.1cm}}
		\newcolumntype{P}[1]{>{\centering\arraybackslash}p{#1}}
		\newcolumntype{L}[1]{>{\raggedright\arraybackslash}p{#1}}
		
		\newcolumntype{A}{
			>{$}r<{$}
			@{\extracolsep{0pt}}
			>{${}} l <{$}
			@{\extracolsep{\fill}}
		}
		
		\usepackage{threeparttable}
		\usepackage{tabularx}
		\newfloatcommand{capbtabbox}{table}[][\FBwidth]
		\newfloatcommand{ImportTable}{input}[][\FBwidth]
		\newfloatcommand{capthreepart}{threeparttable}[][\FBwidth]
		\DeclareFloatFont{Small}{\fontsize{9}{11}\selectfont}
		\DeclareFloatFont{Small2}{\small}
		\DeclareFloatFont{footnote}{\footnotesize}
		\DeclareFloatFont{scriptsize}{\scriptsize}
		\DeclareMarginSet{hangboth}{\setfloatmargins*{\hskip-4cm}{\hskip-4cm}}
		
		\DeclareNewFloatType{figurea}
		{placement=htbp, name=Figure, within=section, fileext=lofa}
		
		\newfloatcommand{ffigabox}{figurea}[\nocapbeside][]
		\makeatletter
		\def\thickhline{%
			\noalign{\ifnum0=`}\fi\hrule \@height \thickarrayrulewidth \futurelet
			\reserved@a\@xthickhline}
		\def\@xthickhline{\ifx\reserved@a\thickhline
			\vskip\doublerulesep
			\vskip-\thickarrayrulewidth
			\fi
			\ifnum0=`{\fi}}
		\makeatother
		
		\newlength{\thickarrayrulewidth}
		\hypersetup{
			colorlinks = true,
			citecolor  = blue,
			urlcolor   = red,
			breaklinks = true
		}

		\newcolumntype{O}{>{\centering\arraybackslash}m{1.5cm}}
		
		\let\origappendix\appendix 
		\renewcommand\appendix{\clearpage\pagenumbering{roman}\origappendix}

		\usepackage{subcaption}
		\usepackage{tablefootnote}

		\usepackage{nicematrix}
		\makeatletter
		\AddToHook{env/threeparttable/begin}
		{\TPT@hookin{NiceTabular}\TPT@hookin{NiceTabular*}}
		\makeatother
		
		\usepackage[final]{pdfcomment}
		
		\usepackage[final]{changes}
		\definechangesauthor[name={Bavo D.C. Campo}, color=red]{B}
		
		\usepackage{bigdelim}
		
		\graphicspath{{Figures/}}
		
		\author[1]{Bavo D.C. Campo}
		\author[1,2,3,4]{Katrien Antonio}
		\affil[1]{Faculty of Economics and Business, KU Leuven, Belgium.}
		\affil[2]{Faculty of Economics and Business, University of Amsterdam, The Netherlands.}
		\affil[3]{LRisk, Leuven Research Center on Insurance and Financial Risk Analysis, KU Leuven, Belgium.}
		\affil[4]{LStat, Leuven Statistics Research Center, KU Leuven, Belgium.}
		\title{\textbf{An engine to simulate insurance fraud network data}}
		
		\allowdisplaybreaks

		\newcommand\Tstrut{\rule{0pt}{3.2ex}}         
		\newcommand\Cstrut{\rule[-1.25ex]{0pt}{0pt}}   

\begin{document}
			\newcommand{\form}[1]{\scalebox{1.087}{\boldmath{#1}}}
			
			\sloppy
			\maketitle
			\begin{abstract}
				\noindent
				Traditionally, the detection of fraudulent insurance claims relies on business rules and expert judgement which makes it a time-consuming and expensive process \citep{PaperMaria}. Consequently, researchers have been examining ways to develop efficient and accurate analytic strategies to flag suspicious claims. Feeding learning methods with features engineered from the social network of parties involved in a claim is a particularly promising strategy (see for example \cite{PaperMaria, van2016gotcha,Tumminello2022}). When developing a fraud detection model, however, we are confronted with several challenges. The uncommon nature of fraud, for example, creates a high class imbalance which complicates the development of well performing analytic classification models. In addition, only a small number of claims are investigated and get a label, which results in a large corpus of unlabeled data. Yet another challenge is the lack of publicly available data. This hinders not only the development of new methods, but also the validation of existing techniques. We therefore design a simulation machine that is engineered to create synthetic data with a network structure and available covariates similar to the real life insurance fraud data set analyzed in \citet{PaperMaria}. Further, the user has control over several data-generating mechanisms. We can specify the total number of policyholders and parties, the desired level of imbalance and the (effect size of the) features in the fraud generating model. As such, the simulation engine enables researchers and practitioners to examine several methodological challenges as well as to test their (development strategy of) insurance fraud detection models in a range of different settings. Moreover, large synthetic data sets can be generated to evaluate the predictive performance of (advanced) machine learning techniques.
				
				\vspace{2mm}
			\end{abstract}

			\input{Introduction.tex}

			\input{FraudOverview.tex}

			\input{SimulationEngine.tex}

			\input{Illustration.tex}
			\input{Discussion.tex}
			
			\input{Acknowledgments.tex}

			\addcontentsline{toc}{chapter}{Bibliography}
			\renewcommand\harvardurl{}
			\bibliographystyle{agsmAdj}
			\bibliography{ReferencesFraud}
			
			\bookmarksetup{startatroot}
			\appendix
			\input{Appendix.tex}

		\end{document}

%% file: Introduction.tex
\section{Introduction}
Fraudulent activity in the insurance industry causes significant financial losses for both insurance companies and policyholders. In non-life insurance, the total yearly cost of fraudulent claims is estimated to be more than \$40 billion in the United States \citep{fbi}.  For an average family, this leads to an increased yearly premium of \$400 to \$700 \citep{fbi}. To detect and mitigate fraud, insurance companies implement various anti-fraud measures. Traditionally, insurance companies rely on a combination of business rules and expert judgment to identify fraudulent claims \citep{PaperMaria}. The business rules flag suspicious claims, which are then sent to experts who determine whether the claim is fraudulent or not \citep{Warren2018}. These in-depth investigations, however, are time-intensive and costly. Researchers have therefore developed insurance fraud detection models combining business rules and analytic\deleted[id=B]{al} techniques to flag the most suspicious claims, which are sent to the experts for further investigation. As such, experts can focus solely on the claims with a high likelihood of fraud and avoid spending precious resources on examining non-fraudulent claims. Insurers predominately rely on analytics to prevent fraud \citep{EIOPA}. Moreover, fraud detection is considered an area for more intense use of big data and analytics in the insurance industry.

Within fraud analytics, researchers rely on a wide range of statistical and machine learning techniques (see \citet{Ngai2011} and \citet{Albashrawi2016} for an overview). In the literature, we find examples of both supervised and unsupervised (machine learning) techniques to construct fraud detection models. \citet{Vosseler2022}, for example, developed an unsupervised anomaly detection technique to identify fraudulent insurance claims. \citet{Nur2020} constructed fraud detection models using neural networks and tree-based machine learning techniques. The accuracy of such models, however, greatly depends on the input. We typically use traditional claim characteristics, such as the claim amount, as features in a fraud detection model \citep{Baesens2015}. These characteristics are static whereas the typical features of fraudsters tend to be dynamic \citep{PaperMaria,Gomes2021, Tumminello2022}. According to \citet{Jensen1997}, fraudsters adapt their tactics in response to fraud detection systems and hence, the typical fraudster profile evolves over time. 

One particularly promising approach to capture the characteristics of fraudsters is through social network analytics \citep{van2016gotcha,PaperMaria, Tumminello2022}. In an insurance context, social network data captures the relationship between claims on the one hand and policyholders and other involved parties (e.g., garage, broker, expert) on the other hand. By analyzing the social network structure of reported claims, insurers can unravel patterns and relationships among policyholders and claims that are indicative of fraud. Moreover, this approach potentially uncovers organized schemes of collaborating fraudsters who are trying to hide their tracks. Criminals often commit crimes in groups to increase rewards and to decrease the risk of detection \citep{Reiss1988,Andresen2009}. Furthermore, in organized crime, such as fraud, social connections play a crucial role since these connections are based on trust and provide access to co-offenders and opportunities \citep{vanKoppen2010}. As such, social network analytics can assist in identifying enduring relationships between fraudsters even as overt characteristics undergo changes.

Nonetheless, publicly available data on insurance fraud is scarce, in particular data with a social network structure. This makes it difficult for researchers to test, validate and improve existing fraud detection methods. Moreover, the lack of data hinders the reproducibility of research findings and the discovery of novel methodologies \citep{Baesens2023}. The main reason for the limited availability is the sensitive nature of the data \citep{Lopez2015}. Insurance data sets contain confidential information about the insurance company and its policyholders. In consequence, the data used to develop and validate fraud detection methods and models is almost never shared. Researchers in fraud analytics are confronted with several inherent methodological challenges when developing analytic models for fraud detection \citep{Baesens2023}. Investigating suspicious claims is a time-intensive and expensive process, which results in only few claims being labeled (i.e., whether the claim is fraudulent or non-fraudulent) \citep{Warren2018}. In addition, due to fraud being uncommon, data sets are often characterized by a severe class imbalance (see, for example \citet{PaperMaria,Gomes2021,Subudhi2020}). Another challenge is the continuous development of machine learning techniques \citep{Baesens2023}. To assess whether these perform better or on par with existing techniques, we need to evaluate competing models or techniques in similar conditions. That is, using the same (type of) data set. 

One way to address the scarcity of publicly available data, is by using a simulation engine to generate synthetic data that mimics the structure of the original data set \citep{Lopez2015}. Simulation engines enable researchers to perform benchmark studies to investigate the properties and performance of various statistical and machine learning techniques \citep{Morris2019,Khondoker2016}. Additionally, synthetic data facilitates the development of new methods and stimulates the reproducibility of research. Recently, simulation engines have gained considerable attention in actuarial science. \citet{Gabrielli2018} developed a simulation engine to generate individual claim histories of non-life insurance claims and \citet{So2021} devised an engine to generate synthetic telematics data. However, both engines employ neural networks trained on a single real insurance data set to generate synthetic data that closely mirrors the original data. Conversely, the simulation machine of \citet{Avanzi2021} does not emulate a single data set when generating individual non-life insurance claims. Instead, it allows users to generate a diverse set of scenarios which vary in complexity.

In this paper, we design a simulation machine that generates synthetic insurance fraud network data. The simulation engine is inspired by and mimics the structure and properties of the non-life motor insurance data used in \citet{PaperMaria}, which contains both traditional claim and policy(holders) characteristics as well as social network features. When generating the synthetic data, the user has control over several data-generating mechanisms. Both policyholder, contract-specific and claim characteristics as well as the dependence between them can be adjusted. Further, when simulating the number of claims, the individual claim costs and the claim labels (i.e., fraudulent or non-fraudulent), the user can specify the (effect size of the) features that are used in the data-generating model. Specific characteristics of the social network structure and fraud investigation process can be adjusted as well. In addition, the size of the resulting data set and the required level of class imbalance can be set by the user. Hereby, the simulation engine provides researchers with a powerful and valuable tool for evaluating and improving the performance of fraud detection methods across various scenarios. The simulation engine's ability to produce large data sets makes it ideal for machine learning techniques that require large amounts of data. Furthermore, researchers and practitioners can use the engine to test their (development strategy of) insurance fraud detection models and to investigate the performance of a wide range of analytic methods in tackling the challenges inherent to fraud data sets, for example the severe class imbalance and large number of missing labels. Additionally, by examining a specific method or model in these scenarios, researchers can gain a better understanding of its strengths and limitations.

This paper is structured as follows. In \Cref{sec:FraudOverview}, we discuss the fraud cycle, existing analytic fraud detection strategies and provide a comprehensive overview of social network analytics for fraud detection strategies. Further, we address several challenges that are an integral part of fraud detection research. \Cref{sec:SimulationEngine} delves into the design of the simulation engine and explains how a synthetic data set is generated. \Cref{sec:Illustration} showcases the simulation engine's capabilities by generating and exploring two different types of synthetic data sets. In the first type we introduce a social network effect when simulating the claim labels and in the second type, we omit the social network effect. Additionally, using the artificial data, we provide a practical demonstration of the development and evaluation of a fraud detection model. We conclude the paper with \Cref{sec:Discussion}.

%% file: OldFiles/FraudOverview.tex
\section{Fighting fraud with data analytics: strategies, techniques and challenges}\label{sec:FraudOverview}
In this section, we provide an overview of some conventional and emerging approaches to fraud detection. We highlight the role of analytics in uncovering fraudulent activities and discuss the various challenges that researchers and practitioners face in this domain.

\subsection{Uncovering fraud: traditional and analytic approaches}\label{subsec:FraudDetection}
In insurance, policyholders file a claim to request a financial compensation for a covered loss. The insurance company retains all relevant information on past and current claims in a database, typically stored in a tabular format. The data set encompasses claim, policyholder, and contract-specific attributes, collectively referred to as traditional claim characteristics. Certain claims, however, are illegitimate and detecting fraudulent claims is essential for insurance companies to prevent financial losses and to protect their policyholders. Hereto, insurers adopt either a traditional, a data-driven or a combined strategy to flag suspicious claims.

\paragraph*{Expert-based fraud detection} The traditional approach to detect fraud is expert-based \citep{Baesens2015}. This approach is two-fold. First, the insurance company flags certain claims as suspicious. To flag suspicious claims, companies rely on a set of business rules that are based on insights from previous investigations. Second, once a claim is flagged as suspicious, the claim is passed to an expert, who conducts an in-depth investigation to determine whether the claim is fraudulent or not \citep{Warren2018}. Hereafter, newly obtained insights from the investigation are used to adjust the procedure for identifying suspicious claims. This is known as the fraud detection cycle.

The expert-based approach, however, has some notable shortcomings \citep{Baesens2015}. It is highly dependent on the manual input and expertise of the expert. In addition, investigating claims is a time-intensive and expensive process. Moreover, the dynamic nature of fraud requires that the rule base to flag suspicious claims needs to be continuously monitored, improved and updated. To address these drawbacks, researchers have developed alternative approaches to detect fraud in a more automated manner \citep{Baesens2015, Ngai2011, Albashrawi2016, Barman2016}. Notwithstanding, even with the alternative approaches, the inclusion of expert knowledge and input is critical to the success of the fraud detection system.

\paragraph*{Fraud analytics} The limitations of the expert-based approach prompted researchers to develop data-driven methodologies to detect fraud. In the literature, either supervised or unsupervised learning techniques or a combination of both are employed. Within fraud detection, we use unsupervised learning techniques to identify anomalous behavior \citep{Baesens2015}. There is a plethora of anomaly detection techniques available and \cite{Hilal2022} provides an extensive overview of anomaly detection techniques to detect financial fraud. In our paper, we focus on supervised learning methods which learn from historical, labeled data. There are numerous supervised techniques available that can be utilized to construct a fraud detection model, ranging from logistic regression models to neural networks. A comprehensive literature review of the supervised techniques applied in financial fraud detection can be found in \cite{Ngai2011,Barman2016,Albashrawi2016}.

One way to tackle insurance fraud detection is by treating it as a binary classification problem. Here, our response variable $Y_i$ can take on only two values: 0 (non-fraud) or 1 (fraud). Further, each claim $i$ has a corresponding covariate vector $\bs{x}_i$. The values herein correspond to a set of features that provide information on the policyholder, contract, claim, network and any other relevant features that can assist in identifying fraudulent claims. The general equation of a predictive classification model is
\begin{equation}\label{eq:FraudClassification}
	\begin{aligned}
		P[Y_{i} = 1 | \bs{x}_{i}] &= f(\bs{x}_{i}) 
	\end{aligned}
\end{equation}
where $P[Y_{i} = 1| \bs{x}_{i}]$ denotes the probability that claim $i$ is fraudulent given covariate vector $\bs{x}_i$ and $f(\bs{x}_{i})$ denotes the predictive model. Logistic regression is one of the most popular models to estimate \eqref{eq:FraudClassification} \citep{Ngai2011,Baesens2015,Barman2016,Albashrawi2016}. Other commonly employed techniques include tree-based learners \citep{Kho2017, Roy2017} and neural networks \citep{Srivastava2016, Ghobadi2016}.

To develop a fraud detection model, we rely on historical, labeled data of past observed fraud behavior \citep{Baesens2015}. This historical data is commonly derived from the expert judgment of previously investigated claims and serves as the foundation for constructing an effective fraud detection model. By training the fraud detection model on labeled claims, we aim to find hidden patterns that allow us to identify new fraudulent claims.

\subsection{Enriching traditional claim characteristics with social network data}\label{subsec:socialnetworkdata}
Both the expert-based and fraud analytics approach commonly rely on traditional claim characteristics stored in a tabular data set. To go beyond this tabular structure, we can rely on social network analytics which extracts information from the relational structure in the data set. As such, we augment the database with supplementary information on the social network structure of the claim and the involved parties. The involved parties are typically the policyholder and the experts involved in the claim \citep{PaperMaria}. Certain contracts involve the active participation of brokers, hereby incorporating them into the network structure. Further, depending on the type of insurance, other parties may be present as well. In motor insurance, for example, we commonly also have the auto repair shop that repaired the vehicle (hereafter referred to as the garage). In constructing the network structure, \citet{PaperMaria} take a holistic view by integrating information across multiple lines of business. In this paper, we focus exclusively on the social network structure within one specific insurance product.

\begin{figure}[!htbp]
	\makebox[\linewidth][c]{%
		\caption{A toy example of a social network in an insurance context.}
		\label{fig:SocialNetwork}
		\begin{subfigure}[t]{.5\textwidth}
			\centering
			\caption{\label{fig:ExampleNetwork}Visualization of the relationships between the claims and the involved parties.}
			\includegraphics[width=.95\textwidth]{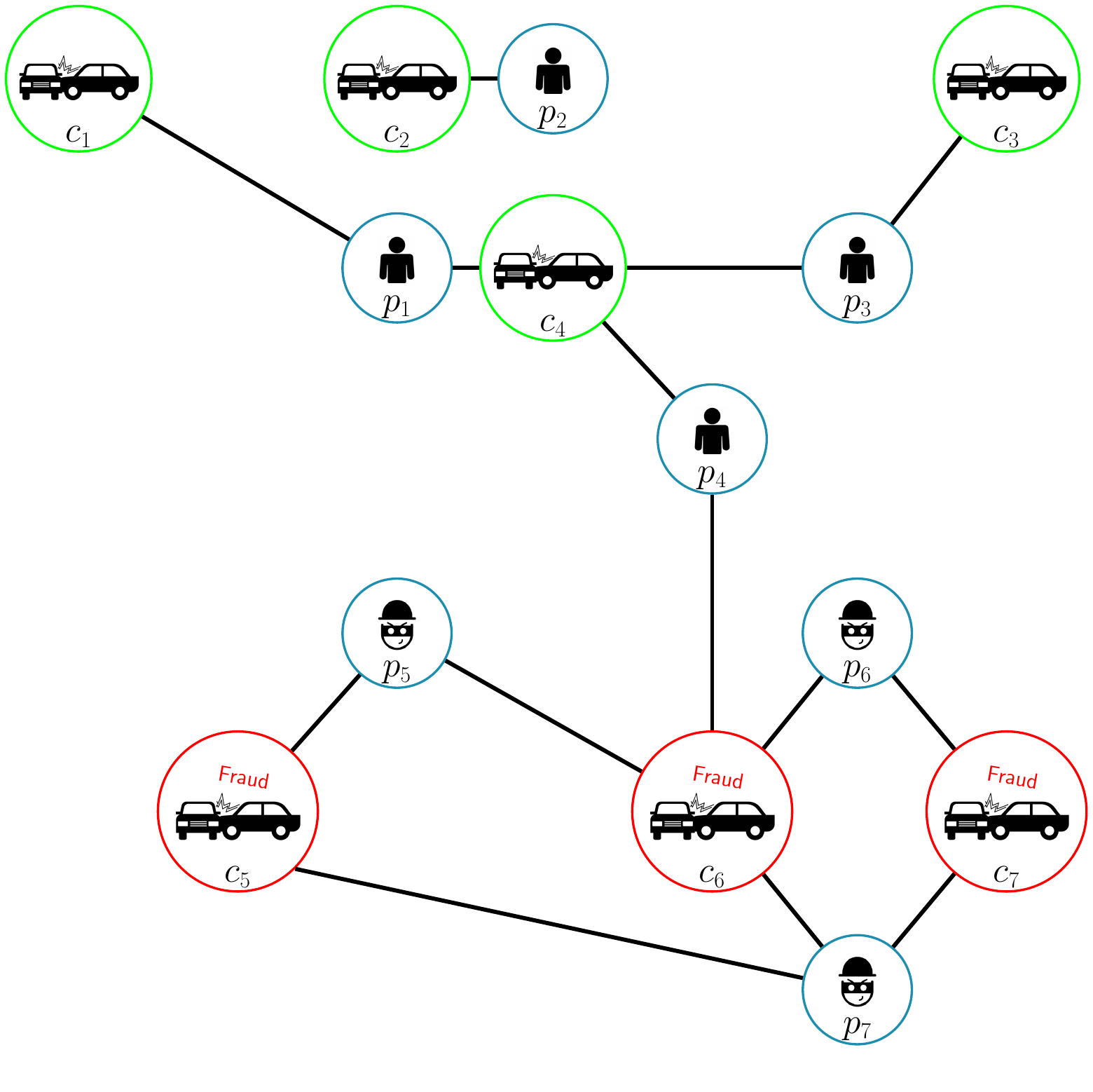}
		\end{subfigure}%
		\hspace{1mm}
		\begin{subfigure}[t]{.5\textwidth}
			\centering
			\caption{\label{fig:WeightMatrix}The weight matrix $\bs{W}$ corresponding to the social network example.}
			\includegraphics[width=.95\textwidth]{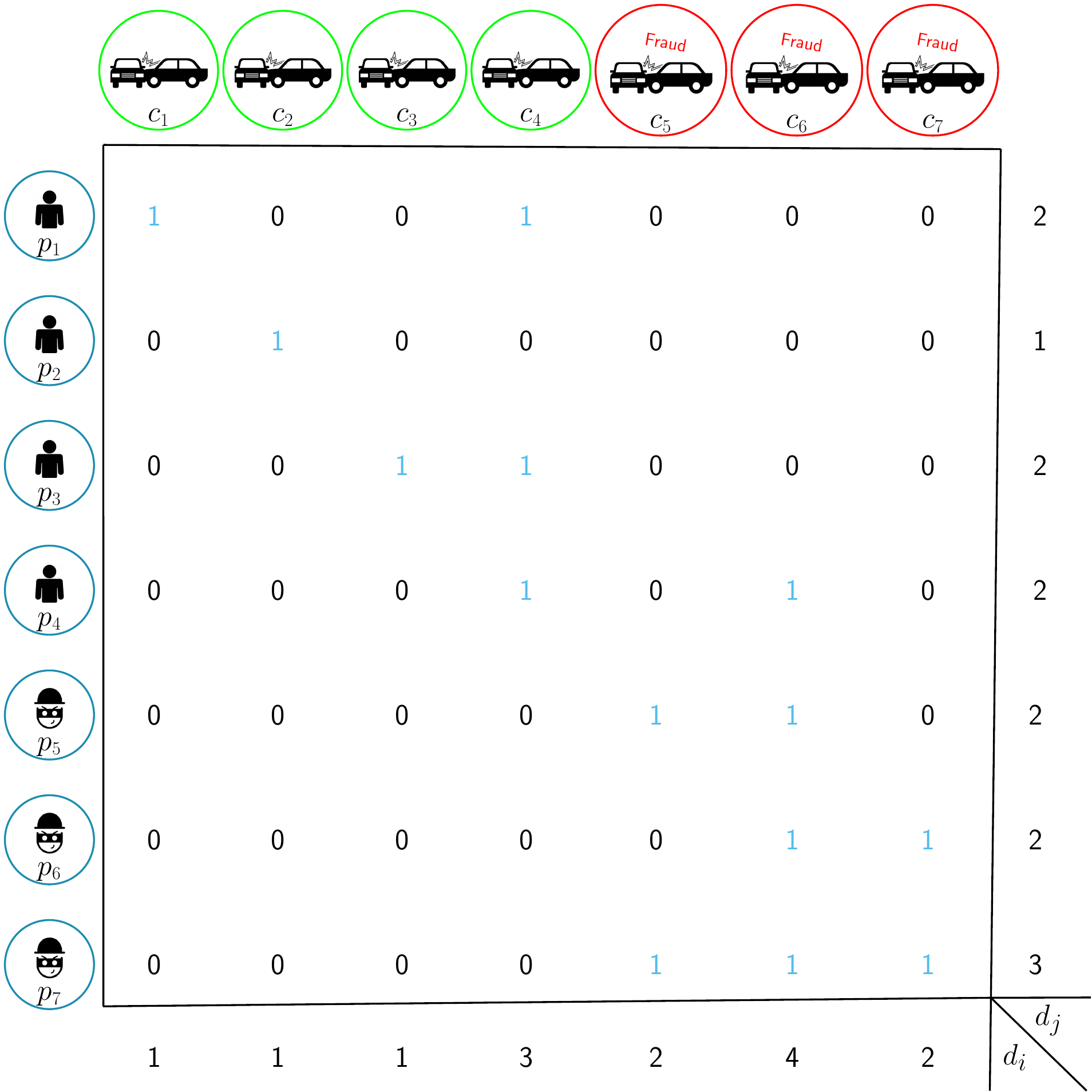}
		\end{subfigure}%
	}\\
\end{figure}

\Cref{fig:SocialNetwork}(a) depicts a toy example of a social network consisting of seven claims and seven involved parties. In this figure, the edges symbolize the connections between the claims and the parties. The claims and parties are represented by circles and the claims in the network are color-coded. Green claims correspond to non-fraudulent claims and fraudulent claims are colored red. There is a notable cluster of fraudulent claims (i.e. $c_5, c_6$ and $c_7$) that are strongly interconnected. Party $p_7$ is connected to all three fraudulent claims and might be the central figure in the criminal network. Via the claims, $p_7$ is connected to the fraudsters $p_5$ and $p_6$. In this example, all fraudsters are connected to each other via one or more fraudulent claims.

To obtain a mathematical representation of the network data, we use a bipartite network of nodes $C \cup P$ and edges $E$. $C$ denotes the set of all claims and $P$ the set of all parties in the network. $E$ is the set of edges that connect the nodes in $C$ to the nodes in $P$. The bipartite network $G = (C \cup P, E)$ is undirected (i.e. there is no direction in the edges). We use $c_i$ to denote an individual claim, where $i \in (1, \dots, n_C)$ and $n_C$ is the total number of nodes in $C$. $p_j$ denotes an individual party. Here, $j \in (1, \dots, n_P)$ where $n_P$ is the total number of nodes in $P$. The network's edges are represented in a weight matrix $\bs{W}$ of dimension $n_P \times n_C$. Each individual edge carries a certain weight $w_{ij}$ that reflects the strength of the relationship between claim $i$ and party $j$. If $w_{ij} > 0$, claim $c_i$ is connected to party $p_j$. We have an unweighted network when all nonzero values in $\bs{W}$ are equal to one. 

We refer to the set of nodes, connected to node $c_i$ via a path of exactly $k$ edges, as the $k^{th}$ order neighborhood of $c_i$ and we denote it as $\mc{N}_{c_i}^k$. Hence, the first-order neighborhood of a claim $c_i$ consists of all involved parties
\begin{equation}
	\begin{aligned}
		\mc{N}_{c_i}^1 = \{p_j | w_{ij} \neq 0\}
	\end{aligned}
\end{equation}
\noindent
and the second-order neighborhood of $c_i$
\begin{equation}
	\begin{aligned}
		\mathcal{N}_{c_i}^2=\{c_k|p_j \in \mathcal{N}_{c_k}^1 \wedge w_{kj}\ne 0\}\setminus c_i.
	\end{aligned}
\end{equation}
are all the claims connected to the parties in $\mc{N}_{c_i}^1$. In an unweighted network, we refer to the number of nodes in a node's first order neighborhood as the degree of the node. For claim $c_i$, we denote this as $d_i$ and $d_j$ refers to the degree of party $p_j$. The degree of all claims is summarized in a $n_C \times n_C$ diagonal matrix $\bs{D}_C$, where $(\bs{D}_C)_{ii} = d_i \ \forall \ i$. Similarly, the $n_P \times n_P$ diagonal matrix $\bs{D}_P$ contains the degrees of all parties.

In this toy example of a social network, we use an unweighted network. \Cref{fig:SocialNetwork}(b) represents the corresponding $\bs{W}$ that captures the connections in the example. The strong interconnectedness between fraudsters and fraudulent claims is also reflected in $\bs{W}$. In the lower right corner of $\bs{W}$ we notice a distinct cluster, which predominantly consists of fraudulent nodes and which reveals a web of fraudsters. We observe another group of connected claims in the upper left corner of $\bs{W}$. This cluster consists mostly of legitimate claims. Furthermore, there are only a few links connecting fraudulent and non-fraudulent nodes, indicating limited relationships between the two groups.

\paragraph*{Homophily} One of the fundamental concepts in a network-based fraud detection approach is the concept of homophily \citep{Baesens2015, PaperMaria}. This refers to the tendency of people to form social connections with individuals that are similar to themselves in some way \citep{Newman2010}. Translated to an insurance context, this means that fraudulent claims are predominantly linked to other fraudulent claims, while non-fraudulent claims tend to be connected to other non-fraudulent claims. Moreover, fraudulent and non-fraudulent claims exhibit a weaker degree of connection with each other.

To assess whether there are patterns of homophily present in the network, we compute the dyadicity and heterophilicity of the network \citep{Park2007, Baesens2015}. Dyadicity measures the connectedness between nodes with the same label. The higher the dyadicity, the more densely connected the same-label nodes are, compared to what is expected based on a random network configuration. Conversely, heterophilicity assesses the degree of interconnection between nodes with different labels. Networks exhibit high heterophilicity when nodes with different labels show higher interconnectedness compared to what is expected by chance.

In a fraud context, the investigated claims can be labeled as fraudulent (1) or non-fraudulent (0). Claims that are uninvestigated have no label and are referred to as unlabeled. Hence, there are three different labels present in the data set. If our focus is on the identification of dense networks of fraudulent claims, we can adopt a one-versus-all classification strategy and group the unlabeled with the non-fraudulent claims. We denote the total number of fraudulent claims as ${}_{1} n_C$ and ${}_{0} n_C$ denotes the total number of non-fraudulent and unlabeled claims. Further, $n_C = {}_{1} n_C +{}_{0} n_C$. In this example, we have three types of relationships between claims or so-called dyads. That is: fraudulent claims connected to other fraudulent claims (1 - 1); fraudulent claims linked to non-fraudulent, unlabeled claims (1 - 0) and non-fraudulent, unlabeled claims connected to non-fraudulent, unlabeled claims (0 - 0). We use $m_{11}$, $m_{10}$ and $m_{00}$ to refer to the total number of dyads of each kind present in the network and $|E| = m_{11} + m_{10} + m_{00}$, where $|E|$ denotes the number of edges. $\rho$ denotes the probability that two nodes are connected and is empirically calculated in the network as
\begin{equation}
	\begin{aligned}
		\rho = \frac{2|E|}{n_C (n_C - 1)}.
	\end{aligned}
\end{equation}
\noindent
If nodes are randomly connected to other nodes irrespective of their labels, the expected values of $m_{11}$ and $m_{10}$ equal \citep{Baesens2015}
\begin{equation}
	\begin{aligned}
		\overline{m}_{11} &= \frac{{}_{1} n_{C} ({}_{1} n_{C} - 1) \rho}{2} \quad \textrm{and} \quad
		\overline{m}_{10} &= {}_{1} n_{C} (n_{C} - {}_{1} n_{C}) \rho.
	\end{aligned}
\end{equation}
\noindent
We then calculate the dyadicity $\mathscr{D}$ and heterophilicity $\mathscr{H}$ of the network as
\begin{align}\label{eq:Homophily}
	\mathscr{D} &= \frac{m_{11}}{\overline{m}_{11}},\\
	\mathscr{H} &= \frac{m_{10}}{\overline{m}_{10}}.
\end{align}
\noindent
When $\mathscr{D} > 1$, the network is dyadic and fraudulent nodes are more densely connected to each other compared to what we expect by chance and $\mathscr{D} \approx 1$ corresponds to a random network configuration. Here, $\approx$ denotes approximately equal to. We have a heterophobic network if $\mathscr{H} < 1$, indicating that fraudulent claims have fewer connections to non-fraudulent claims than what is expected by chance. In a random network configuration, $\mathscr{H} \approx 1$. In our fictive example depicted in \Cref{fig:SocialNetwork}, the network is dyadic ($\mathscr{D} = 2.5$) and heterophobic ($\mathscr{H} = 0.28$). Consequently, we can infer that our network exhibits homophily as it displays both dyadicity ($\mathscr{D} > 1$) and heterophobia ($\mathscr{H} < 1$). In this toy example, engineering features from the social network potentially enables us to identify collaborating fraudsters that try to hide their tracks.

\paragraph*{BiRank algorithm} In a homophilic network (i.e. where $\mathscr{D} > 1$ and $\mathscr{H} < 1$), we can potentially uncover fraud by inspecting claims that are closer and more densely connected to known fraudulent claims. To evaluate the proximity to fraudulent claims, a suitable metric is needed. One effective approach is to employ the BiRank algorithm \citep{BiRank}. This algorithm is an extension of the personalized PageRank algorithm \citep{Page1999} and is specifically designed for bipartite networks. Using BiRank, we rank claims with respect to their exposure to known fraudulent claims \citep{PaperMaria}. 

The scores of nodes $c_i$ and $p_j$ are calculated iteratively as
\[
c_i=\sum_{j=1}^{n_P}w_{ij}p_j\quad \textrm{and} \quad p_j=\sum_{i=1}^{n_C}w_{ij}c_i
\] 
where $c_i$ and $p_j$ denote the scores of nodes $c_i$ and $p_j$, respectively. To ensure convergence and stability, the scores are normalized using
\begin{equation}\label{eq:rank}
	c_i=\sum_{j=1}^{n_P}\frac{w_{ij}}{\sqrt{d_i}\sqrt{d_j}}p_j\quad \textrm{and} \quad p_j=\sum_{i=1}^{n_C}\frac{w_{ij}}{\sqrt{d_i}\sqrt{d_j}}c_i.
\end{equation}
This normalization lessens the contribution of high-degree nodes and gives better quality results \citep{BiRank}. To steer the scoring towards fraudulent claims, we incorporate query vectors in the scoring process. These query vectors encode our prior belief on the nodes' importance. We use $\bs{c}^0$ and $\bs{p}^0$ to denote the query vectors for the claims and parties, respectively. Further, $c^0_i$ and $p^0_j$ represent the individual entries in the vectors $\bs{c}^0$ and $\bs{p}^0$. We adjust \eqref{eq:rank} to
\begin{equation}\label{eq:BiRank1}
	c_i=\alpha \sum_{j=1}^{n_P}\frac{w_{ij}}{\sqrt{d_i}\sqrt{d_j}}p_j + (1-\alpha)c_i^0\quad \textrm{and} \quad p_j=\beta\sum_{i=1}^{n_C}\frac{w_{ij}}{\sqrt{d_i}\sqrt{d_j}}c_i + (1-\beta)p_j^0.
\end{equation}
where $\alpha\in[0,1]$ and $\beta\in[0,1]$ are adjustable parameters. The values for $\alpha$ and $\beta$ regulate the relative emphasis given to the network structure and the query vector. We rewrite \eqref{eq:BiRank1} in matrix form
\begin{equation}
	\label{eq:BiRank2}
	\mathbf{c}=\alpha \boldsymbol{S} \mathbf{p} +(1-\alpha)\mathbf{c}^0\quad \textrm{and} \quad \mathbf{p}=\beta \boldsymbol{S}^T \mathbf{c} +(1-\beta)\mathbf{p}^0.
\end{equation}
Here, $\boldsymbol{S}=\boldsymbol{D}_C^{-\frac{1}{2}}\boldsymbol{W} \boldsymbol{D}_P^{-\frac{1}{2}}$ denotes the symmetrically normalized weight matrix. We start the algorithm by randomly initializing the ranking vectors $\bs{c}$ and $\bs{p}$. Hereafter, we iteratively compute the node scores until convergence.

We encode information about known fraudulent claims into the query vector $\bs{c}^0$ to rank the nodes' scores towards fraudulent claims. When the claim is fraudulent, we set $\bs{c}_i^0 = 1$ and $\bs{c}_i^0 = 0$ otherwise. We define $\mathbf{p}^0\equiv \mathbf{0}$, since only claims can be fraudulent and not parties. We set $\beta=1$ since we do not include prior information on the parties. We adjust \eqref{eq:BiRank2} to
\[
\mathbf{c}=\alpha S \mathbf{p} +(1-\alpha)\mathbf{c}^0\quad \textrm{and} \quad \mathbf{p}= S^T \mathbf{c}.
\]
The iterative procedure to compute the fraud scores is summarized in Algorithm \ref{alg1}. The BiRank algorithm stops when the difference between two successive iterations is below a certain threshold or when we exceed the maximum number of iterations.

\IncMargin{1em}
\begin{algorithm} 
	\caption{BiRank algorithm for computing fraud scores in a network of insurance claims and parties \citep{PaperMaria}. Adapted from Algorithm 1 in \citet{BiRank}. We omit the query vector $\mathbf{p}^0$ and set $\beta=1$.} 
	\label{alg1} 
	\KwIn{Weight matrix $\bs{W}$, query vector $\mathbf{c}^0$ and hyperparameter $\alpha = 0.85$;}
	\KwOut{Ranking vectors  $\mathbf{c}$ and $\mathbf{p}$;}
	Symmetrically normalize $\boldsymbol{W}$: $\boldsymbol{S}=\boldsymbol{D}_P^{-\frac{1}{2}}\boldsymbol{W} \boldsymbol{D}_C^{-\frac{1}{2}}$\;
	Randomly initialize $\mathbf{c}$ and $\mathbf{p}$\;
	\While{stopping criteria is not met}
	{
		$\mathbf{c}\leftarrow\alpha S \mathbf{p} +(1-\alpha)\mathbf{c}^0$\;
		$\mathbf{p}\leftarrow \boldsymbol{S}^T \mathbf{c}$ \;
	}
	\Return{ $\mathbf{c}$ and $\mathbf{p}$};\
\end{algorithm}

\paragraph*{Network featurization} Next, we engineer several social network features from the network structure, the labels and the scores resulting from the BiRank algorithm. These features capture the information that is embedded in the network of the claims and can be integrated into a tabular dataset alongside the individual characteristics of each claim. The social network features then represent an additional source of information that we can use in our fraud analytics models (see \Cref{subsec:FraudDetection}). We divide the network features into two groups, the fraud-score based features and the neighborhood based features (see \Cref{tab:SocialNetworkFeatures}). The results from the BiRank algorithm are used to compute the fraud-score based features. Here, we look at the claim's fraud score and the distribution of the fraud scores in, for instance, its first and second order neighborhood. To summarize these distributions, we can rely on (robust) central tendency measures such as the median or midmean \citep{Tukey1977}. Further, we engineer neighborhood based features that capture the surrounding network structure of each claim. Here, we can compute the size of a claim's first and second neighborhood for instance.

\begin{table}[!htbp]
	\centering
	\caption{\label{tab:SocialNetworkFeatures}Fraud-score and neighborhood based features, partially based on the feature engineering process from \citet{PaperMaria}.}
	\small
	\begin{tabular}{@{\extracolsep{1pt}}llcL{10.25cm}@{}}
		\toprule
		&Name&Order&Description\\
		\midrule
		\multirow{13}{*}{\rotatebox{90}{Fraud-score}}
		&\texttt{scores0}&0&The node's fraud score as determined via BiRank\\
		&\texttt{n1.q1}&1&The first quartile of the empirical distribution of the fraud scores in the node's first order neighborhood\\
		&\texttt{n1.med}&1&The median of the empirical distribution of the fraud scores in the node's first order neighborhood\\
		&\texttt{n1.midmean}&1&The midmean (or interquartile mean) of the empirical distribution of the fraud scores in the node's first order neighborhood\\
		&\texttt{n2.q1}&2&The first quartile of the empirical distribution of the fraud scores in the node's second order neighborhood\\
		&\texttt{n2.med}&2&The median of the empirical distribution of the fraud scores in the node's second order neighborhood\\
		&\texttt{n2.midmean}&2&The  midmean (or interquartile mean) of the empirical distribution of the fraud scores in the node's second order neighborhood\\
		\midrule
		\multirow{8}{*}{\rotatebox{90}{Neighborhood}}
		&\texttt{n1.size}&1&The number of nodes in the node's first order neighborhood\\
		&\texttt{n2.size}&2&The number of nodes in the node's second order neighborhood\\
		&\texttt{n2.ratioFraud}&2&The number of known fraudulent claims in the node's second order neighborhood divided by \texttt{n2.size}\\
		&\texttt{n2.ratioNonFraud}&2&The number of known non-fraudulent claims in the node's second order neighborhood divided by \texttt{n2.size}\\
		&\texttt{n2.binFraud}&2&A binary value indicating whether there is a known fraudulent claim in the node's second order neighborhood\\
		\bottomrule
	\end{tabular}
\end{table}

\subsection{Challenges within fraud analytics}\label{subsec:challenges}
When developing an analytic model for fraud detection, we encounter several challenges that are inherent to research on fraud. One of the main challenges is the infrequent nature of fraud, which leads to highly imbalanced data sets \citep{Baesens2015, Jensen1997, West2016, Thabtah2020}. This imbalance creates a bias towards the majority class for certain analytic techniques, resulting in compromised fraud detection performance. Within fraud research, we commonly tackle the class imbalance problem by employing resampling techniques such as under- and over-sampling, SMOTE or ROSE \citep{Baesens2023, Subudhi2020,Sundarkumar2015,van2016gotcha, PaperMaria}. A second challenge is the dynamic nature of fraud \citep{Baesens2015, Baesens2023, West2016}. To remain undetected, fraudsters continuously adapt their behavior and tactics. Consequently, it is essential to detect fraud as soon as possible and to tweak or rebuild the fraud detection model when necessary. Hereto related is the computational efficiency of the fraud detection model \citep{Baesens2015, West2016} which represents yet another challenge. With the rapid advancement of machine learning, new methods are constantly emerging, adding to the ever-growing repertoire of techniques available. Further, given the high cost of fraud, it is crucial to detect fraudulent activities instantly. Existing analytic methods for fraud detection are predominantly evaluated based on their accuracy, often overlooking the aspect of computational efficiency. For an accurate and reliable comparison of competing methods, it is essential to evaluate their performance on a set of benchmark data sets. Most studies, however, do not share their data sets due to their sensitive nature. The lack of publicly available data hinders the reproducibility of research \citep{Baesens2023} and gives rise to a fifth challenge \citep{Pourhabibi2020, West2016}. The sixth and final challenge arises from the disproportionate misclassification cost and the specific performance criteria to evaluate the model \citep{West2016, Baesens2023}. Analytic fraud models are commonly assessed using performance measures that evaluate the predictive performance. Notwithstanding, wrongly classifying claims has financial implications. Depending on the allocated budget for the fraud investigation process, wrongly classifying a fraudulent claim as legitimate can be considerably more expensive than the reverse. Consequently, it might be more appropriate to compare models in terms of their monetary performance \citep{Baesens2023}.


%% file: OldFiles/SimulationEngine.tex
\section{Simulation engine}\label{sec:SimulationEngine}
Our proposed simulation engine addresses one of the key challenges within fraud research. That is, the limited availability of publicly accessible data.  In this section, we provide an in-depth overview of the data generation process and the architecture of the simulation engine. We outline the sequential steps taken to generate a realistic and representative insurance fraud network data set.

The resulting synthetic data set is structured in a tabular format, with each row representing a unique claim and its corresponding attributes. The columns in the data set capture items such as policyholder characteristics, traditional claim features, involved parties and social network features. These are the attributes typically used for fraud analytics. Further, for each claim we have two different types of labels. The first is the true label of the claim, indicating if it is fraudulent or not. This label is determined by the fraud generating model. The second type of label is the outcome of the fraud investigation, which can either take on the value fraudulent, non-fraudulent or uninvestigated. This variable is typically the one we have available in insurance fraud data sets.

\paragraph*{Architecture} We generate a synthetic, tabular data set in seven consecutive steps (see \Cref{fig:Roadmap}). We start by generating the policyholder characteristics (step 1). Hereafter, we simulate the contract-specific attributes per policyholder (step 2). We use the policyholder and contract-specific features as input for our data-generating claim frequency model and generate the number of claims per contract (step 3). Next, we simulate the individual claim amounts using a data-generating claim severity model (step 4). Similarly to step 3, we use the policyholder and contract-specific characteristics as input for the data-generating model. Subsequently, we combine all simulated claims and their characteristics into a tabular data set and we proceed with the observations that have at least one claim. In step 5, we generate the network structure of the claims by connecting each claim to different types of parties. Next, we engineer the social network features and generate the claim labels which represent the ground truth of the claim (i.e. fraudulent or non-fraudulent). We replicate the fraud investigation process in step 6. Hereby, we generate the label that is commonly available in fraud data sets (see \Cref{sec:FraudOverview}). This label expresses whether the claim has been investigated for fraud and what the outcome of that investigation was. We conclude the synthetic data generation with step 7 where we merge all simulated data.

The simulation engine offers a range of customizable features. In all seven steps, several parameters can be adjusted. Moreover, it allows for dependencies between certain features. For example, we can specify a negative dependence between the age and the value of the car. Consequently, older cars will be characterized by a lower value. By allowing for dependencies between features, the synthetic data captures more realistic and nuanced relationships among variables and more closely mirrors real-world data sets. Following, we discuss the seven consecutive steps in detail. An extensive overview of the default configuration is described in \cref{app:DefaultConfiguration}.
\afterpage{%
	\begin{landscape}
		\null\vfill
		\begin{figure}[!htbp]
			\centering
			\caption{\label{fig:Roadmap}Roadmap of the simulation engine.}
			\makebox[\textwidth][c]{\includegraphics[width = \textwidth]{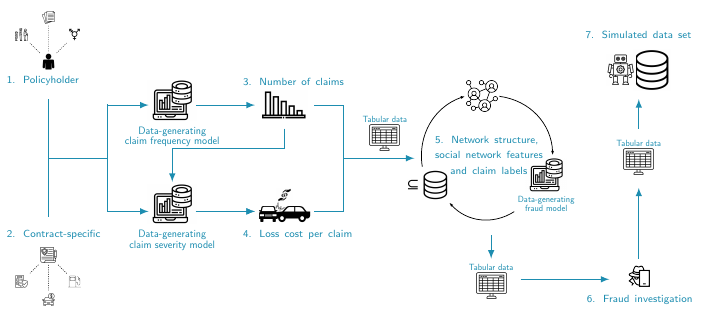}}
		\end{figure}
		\vfill
	\end{landscape}
}

\afterpage{%
	\begin{landscape}
		\begin{table}
			\caption{The policyholder and contract-specific characteristics, along with the generator used to simulate the feature values.}
			\label{tab:PHContractCharacteristics}
			\small
			\begin{threeparttable}
			\begin{tabular}{@{\extracolsep{1pt}}llcL{12cm}L{6.5cm}@{}}
				\toprule
				&Variable & Type & Description & Generator\\
				\midrule
				\multirow{8}{*}{\rotatebox{90}{Policyholder}}
				&\texttt{IDPH} & Continuous & Unique ID to identify the policyholder (index $i = 1, \dots, n_{ph}$) &\\
				&\texttt{AgePH} & Continuous & Age of the policyholder in years. Default range is [18, 80] & $\mathcal{N}(40, 15)$ \\
				&\texttt{GenderPH} & Categorical & Gender of the policyholder: \texttt{female} ($u_i \leq 0.28$), \texttt{male} ($u_i > 0.29$) or \texttt{non-binary} ($0.28 \leq u_i \leq 0.29$) & $u_i \sim U(0, 1)$  \\
				&\texttt{ExpPH} & Continuous & Time since inception of the first contract of the policyholder, in years & $\mathcal{N}(5, 1.5)$ \\
				&\texttt{RateNrContracts} & Continuous & Rate parameter $\lambda_i$ for generating the number of contracts &  $\lambda_i = 0.25 (1.05 - 2.5 \times 10^{-6} \times \text{\texttt{AgePH}}_i + 0.0025 \times \text{\texttt{AgePH}}_i^2 - 2.65 \times 10^{-5} \times \text{\texttt{AgePH}}_i^3)$\\
				&\texttt{NrContractsPH} & Ordinal & Number of contracts. Default range is [1, 5]& $\text{Poi}(\lambda_i)$\\
				\midrule
				
				\multirow{14}{*}{\rotatebox{90}{Contract-specific}}
				&\texttt{ContractID} & Continuous & Unique ID to identify the contract (index $j = 1, \dots,$ $\text{\texttt{NrContractsPH}}_{i}$)&\\
				&\texttt{ExpPHContracts} & Continuous & Duration or exposure of the contract, in years & if \texttt{NrContractsPH}$_{i} > 1$:\\
				&&&& \hspace{0.5em}\texttt{ExpPH}$_{i} - $ U$(0, \text{\texttt{ExpPH}}_{i} / 2)$\\
				&&&& else:\\
				&&&& \hspace{0.5em} \texttt{ExpPHContracts}$_{ij} = $ \texttt{ExpPH}$_{i}$\\
				&\texttt{AgeCar} & Continuous & Age of the vehicle in years & $\max(\mathcal{N}(7.5, \sqrt{5}), \text{\texttt{ExpPHContracts}}_{ij})$\\
				&\texttt{OrigValueCar} & Continuous & Original value of the vehicle & $\text{Exp}(\lambda_i / \texttt{NrContractsPH}_i) \upsilon$ \\
				&\texttt{ValueCar} & Continuous & Current value of the car & $\text{\texttt{OrigValueCar}}_{ij} (1 - \delta)^{\text{\texttt{AgeCar}}_{ij}}$\\
				&\texttt{Coverage} & Categorical & Type of coverage provided by the insurance company: & $\text{Multinomial}(1, \pi_{\text{TPL}}, \pi_{\text{PO}}, \pi_{\text{FO}})$\\
				&&& \hspace{0.5em}\texttt{TPL} = only third party liability,& (see Appendix \ref{app:Coverage})\\
				&&& \hspace{0.5em}\texttt{PO} = partial omnium = TPL + limited material damage,&\\
				&&& \hspace{0.5em}\texttt{FO} = full omnium = TPL + comprehensive material damage.&\\
				&\texttt{Fuel} & Categorical & Type of fuel of the vehicle: \texttt{Gasoline/LPG/Other} (0) or \texttt{Diesel} (1) & Bernoulli$(0.3)$\\
				&\texttt{BonusMalus} & Ordinal & Level occupied in bonus-malus scale of the insurance company & $\lfloor \mc{G}(1, 1 / 3) \rfloor$ \\
				\midrule
				
				\multirow{6}{*}{\rotatebox{90}{Claim}}
				& \texttt{ClaimAge} & Integer & Number of months from beginning of contract to the date of the incident & $\min(\lfloor \text{Exp}(0.25) \rfloor, \lfloor \text{\texttt{ExpPHContracts}}_{ij} * 12 \rfloor)$\\
				& \texttt{ClaimDate} & Continuous & Number of years between the start of the contract and the & $\max(U(0, \text{\texttt{ExpPHContracts}}_{ij}),$\\
				&&& claim's filing date & \hspace{1.75em} $\texttt{ClaimAge}_{ijk} / 12)$\\
				&\texttt{Police} & Categorical & Whether police was called when the incident happened: \texttt{no} (0) or \texttt{yes} (1) & Bernoulli$(0.25)$\\
				&\texttt{nPersons} & Integer & Number of other persons involved in the claim (see \Cref{subsec:Fraud}). Range is [0, 5] &  $S \xleftarrow{\pi_p} x$\\
				
				\bottomrule
			\end{tabular}
			\begin{tablenotes}
			\item[] $\mathcal{N}$ denotes the normal distribution, $U$ the uniform distribution, Poi the Poisson distribution, $\mc{G}$ the Gamma distribution and Exp denotes the exponential distribution. $\upsilon = 25 \times 10^3$ if \texttt{GenderPH}$_i$ = \texttt{male}, $\upsilon = 20 \times 10^3$ if \texttt{GenderPH}$_i$ = \texttt{female} and $\upsilon = 22.5 \times 10^3$ if \texttt{GenderPH}$_i$ = \texttt{non-binary}. $\delta$ is the depreciation rate (see \Cref{tab:Dependencies}). We use $\lfloor \cdot \rfloor$ to depict the floor function. $S \xleftarrow{\pi_p} x$ denotes that a random sample $x$ is drawn from the set $S = (0, 1, 2, 3, 4, 5)$ with corresponding probability $\pi_p = (0.025, 0.6, 0.2, 0.1, 0.1, 0.025)$.
			\end{tablenotes}
			\end{threeparttable}
		\end{table}
	\end{landscape}
}

\subsection{Policyholder and contract-specific characteristics}\label{subsec:PHConstractCharacteristics}
Prior to the synthetic data generation, the user can adjust several parameters that determine the characteristics of the synthetic data (see \cref{app:DefaultConfiguration} for the default configuration). Hence, the user has full control over the data-generating mechanics. One of the adjustable parameters is the number of policyholders $n_{ph}$, which determines the size of the resulting data set. By default, $n_{ph} = 10 \ 000$. We use $i = (1, \dots, n_{ph})$ as an index for the policyholders. In addition to $n_{ph}$, the user can also specify the total number of experts $n_{exp}$, garages $n_{gar}$, brokers $n_{bro}$ and other person(s) involved in the claim $n_{per}$. These parameter values will govern the size of the social network.

We start the synthetic data generation by simulating the policyholder characteristics for $n_{ph}$ policyholders (step 1 in \Cref{fig:Roadmap}). Here, we generate features such as the age and gender of the policyholder and the number of contracts. We use \texttt{NrContractsPH}$_i$ to denote the number of contracts of policyholder $i$ and use $j = (1, \dots, \text{\texttt{NrContractsPH}}_{i})$ as an index for the contracts. Per policyholder, we also generate the number of years since the inception of the first contract and refer hereto as the exposure $w_i$. By default, we set the average exposure to five years and the maximum to 20 years. Additionally, we simulate the contract-specific exposure $w_{ij}$. In the synthetic data set, we consolidate the multiple years of coverage into a single contract to simplify the data structure and analysis. That is, by default we allow $w_{ij} > 1$. \Cref{tab:PHContractCharacteristics} gives an overview of the different attributes that are generated. The first column in this table depicts the variable name, the second the variable type and the third column contains the feature description. The last column of \Cref{tab:PHContractCharacteristics} specifies which generator is used to simulate the feature values. For certain features, the user can specify the range of the feature values. When generating the data, we ensure that all simulated values fall within this prespecified range (see \cref{app:LimitRange}). Hereby, we avoid generating implausible or invalid values. For the policyholder's age, for example, the default range is [18, 80]. Consequently, none of the policyholders will be younger than 18 (the legal driving age in many countries) or older than 80. Once all policyholder characteristics are generated, we proceed to simulate the contract-specific features such as the age of the car and the type of coverage (see \Cref{tab:PHContractCharacteristics}). 

\paragraph*{Dependence structure} The simulation engine allows to specify a dependency between different features. To generate the dependency, we rely on copulas \citep{Denuit2005}. In our simulation engine, we restrict ourselves to the bivariate Ali-Mikhail-Haq (AMH) \citep{Kumar2010} and Frank copula \citep{Denuit2005}. We use $\theta$ to denote the parameter that controls the dependence.

\Cref{tab:Dependencies} presents an overview of the dependencies and the method used to incorporate them. Within insurance, we commonly have variables that are correlated \citep{Goldburd2016}. Consequently, by allowing for dependencies, we can create a more realistic data set. For example, in real life data sets we commonly observe that older cars are worth less compared to newer cars. In our engine, we incorporate this negative dependency between the age of the car and its value by using a Frank copula with $\theta = -25$.

\begin{table}[!htbp]
	\centering
	\caption{Overview of the dependencies between the variables.}
	\label{tab:Dependencies}
	\small
	\begin{tabular}{lL{8.75cm}}
		\hline
		Variables & Dependency\\
		\hline
		\texttt{AgePH} and \texttt{GenderPH} & Weak negative dependence, introduced using AMH copula with $\theta = -0.15$\\
		\texttt{AgePH} and \texttt{ExpPH} & Weak positive dependence, introduced using AMH copula with $\theta = 0.15$\\
		\texttt{AgePH} and \texttt{NrContracts} & Convex function (see \Cref{tab:PHContractCharacteristics}) and positive dependence between \texttt{AgePH} and \texttt{NrContracts}, introduced using AMH copula with $\theta = 0.95$\\
		\texttt{AgeCar} and \texttt{OrigValueCar} & Negative dependence, introduced using Frank copula with $\theta = -25$\\
		\texttt{OrigValueCar} and \texttt{ValueCar} & The depreciation rate $\delta = 0.15$ for cars whose original value $< 30 \ 000$ and $\delta = 0.075$ when the original value $\geq 30 \ 000$ (see \Cref{tab:PHContractCharacteristics} and \citet{CarDepreciation}).\\
		\texttt{Coverage} and \texttt{ValueCar}, \texttt{AgeCar}, \texttt{AgePH} & A dependency is introduced using a multinomial logistic regression model (see Appendix \ref{app:Coverage})\\
		\hline
	\end{tabular}
\end{table}

\subsection{Claim frequency and claim severity}\label{subsec:ClaimSim}
Next, we proceed to simulating the number of claims $N_{ij}$ for policyholder $i$ on contract $j$ and the individual claim costs $L_{ijk}$. We use $k = (1, \dots, N_{ij})$ as an index for the claims. Hereto, we employ the frequency-severity approach \citep{Ohlsson,Frees2014} where we model the claim frequency and claim severity separately. To simulate the number of claims and the claim costs as a function of the policyholder and contract-specific characteristics, we rely on the generalized linear model (GLM) framework \citep{GLM}. We use a Poisson GLM with log link as the data-generating model for the number of claims \citep{Ohlsson,Xacur2015}
\begin{equation}\label{eq:ClaimFrequency}
	\begin{aligned}
		N_{ij} \sim \text{Poi}(w_{ij} \exp({}_{cf} \bs{x}_{ij}^\top \bs{\beta}_{cf})).
	\end{aligned}
\end{equation}
To generate $N_{ij}$, we take a random draw from $\text{Poi}(w_{ij} \exp({}_{cf} \bs{x}_{ij}^\top \bs{\beta}_{cf}))$, with ${}_{cf} \bs{x}_{ij}$ the covariate vector and $\bs{\beta}_{cf}$ is the corresponding parameter vector. The exposure $w_{ij}$ of the contract is included as an offset term and the subscript $cf$ stands for claim frequency. In our simulation engine, the user can specify which features should be included in ${}_{cf} \bs{x}_{ij}$ and the features' effect size can be adjusted via $\bs{\beta}_{cf}$. As such, the user can control the relation between the $N_{ij}$'s and the policyholder and contract-specific characteristics (see Appendix \ref{app:ClaimFreqSev} for the default specification of the claim frequency model).

Hereafter, we generate the claim-specific characteristics (see \Cref{tab:PHContractCharacteristics}). For example, we simulate the duration in months since the beginning of the contract until the date of the incident. Hereby, we create the claim-specific information that is typically available in fraud insurance data sets and that is used in fraud detection models (see, for example, \citet{PaperMaria}).

Next, we proceed with generating the cost $L_{ijk}$ of claim $k$ under contract $j$ of policyholder $i$. The data-generating model for the claim amounts $L_{ijk}$ is driven by a gamma GLM with log link
\begin{equation}\label{eq:ClaimSeverity}
	\begin{aligned}
		L_{ijk} \sim \mathcal{G}(\alpha, \alpha / \exp(\bs{x}_{ij}^\top \bs{\beta} + N_{ij} \zeta))
	\end{aligned}
\end{equation}
\noindent
where $\mathcal{G}$ denotes the gamma\footnote{For the gamma distribution, we use the parameterization with the density function $f(x) = \tau^\alpha x^{\alpha - 1} \exp(-\tau x) / \Gamma(\alpha)$ where $\tau_i = \alpha / \exp(\bs{x}_{ij}^\top \bs{\beta} + N_{ij} \zeta)$ and $\Gamma(\cdot)$ denotes the gamma function.} distribution and $\alpha$ the shape parameter, which we set to $0.25$. The subscript $cs$ stands for claim severity and the parameter $\zeta$ controls the dependency between the claim frequency and claim severity \citep{Frees2011, Garrido2016}. Within the frequency-severity approach, we commonly assume that $\zeta = 0$ (i.e. the claim frequency and claim severity are independent). We specify 50 as a lower limit for $L_{ijk}$ to prevent generating implausible low claim amounts. Consequently, when $L_{ijk} < 50$ we replace it with a randomly drawn value from $U(50, 150)$ to ensure that a low yet realistic claim amount is generated. As with the data-generating claim frequency model (see equation \eqref{eq:ClaimFrequency}), we can specify which features are included in ${}_{cs}\bs{x}_{ij}$ as well as their effect size via $\bs{\beta}_{cs}$ (see Appendix \ref{app:ClaimFreqSev} for the default model specification).

\subsection{Constructing the social network structure and simulating fraudulent claims}\label{subsec:Fraud}
\begin{figure}[!ht]
	\centering
	\caption{\label{fig:exampleSocialNetwork}Example of a social network structure, which illustrates the desired connectivity we want to obtain in our synthetic data set. Each claim is linked to specific parties, and as a result, claims that share the same party are connected to each other in the network. The rectangles depict the parties and the circles the claim. Red claims are fraudulent claims, green claims legitimate claims and the gray claims represent unlabeled claims.}
  	\includegraphics[width = \textwidth]{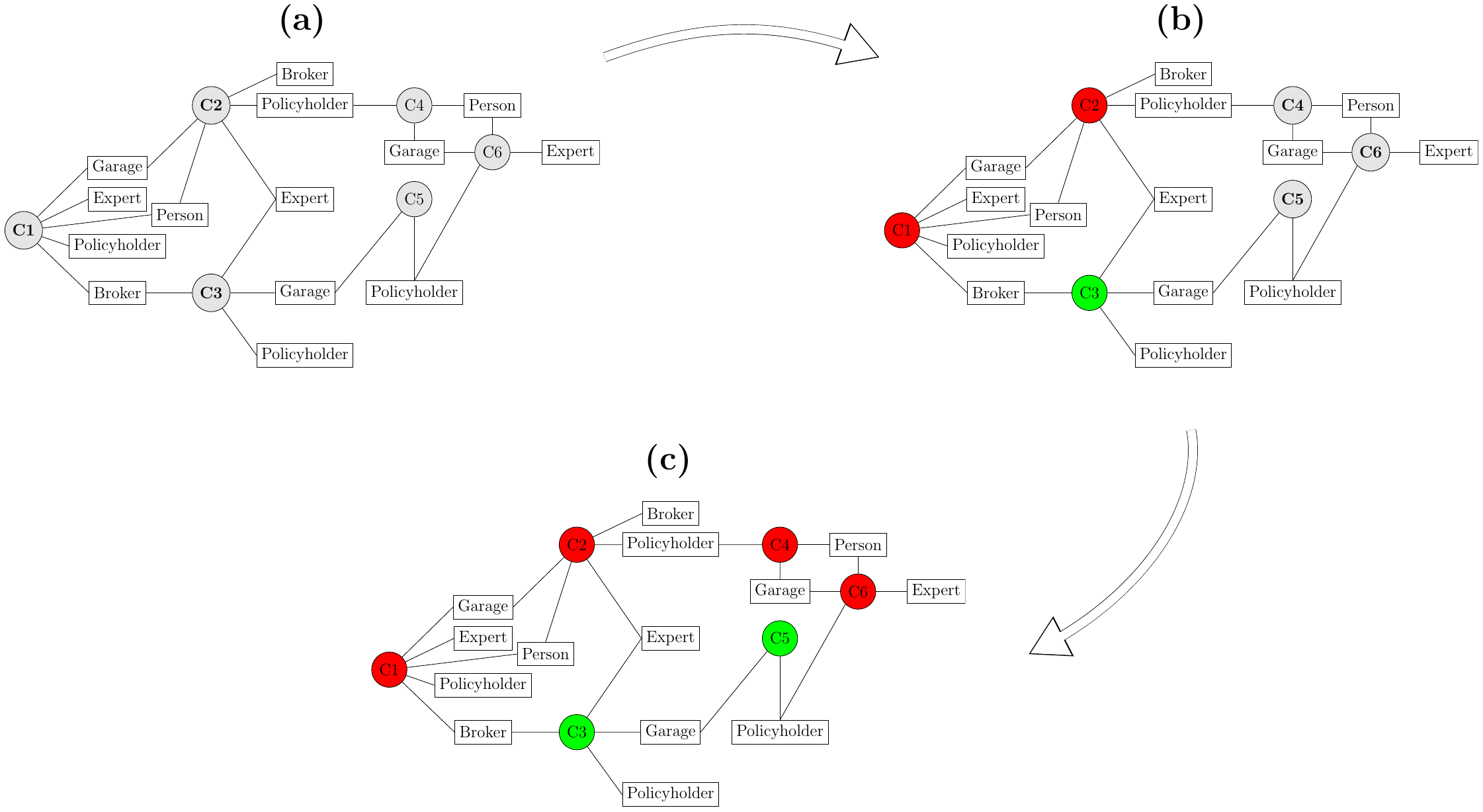}
\end{figure}

Our next objective is to generate the social network structure that links claims to parties and to other claims. Within motor insurance each claim is linked to a policyholder and a garage. Other parties involved in the claim may include brokers and persons other than the policyholder. Experts are involved in the process only when the claim amount exceeds a certain threshold \citep{KBCBrussels}. Insurance companies commonly handle minor losses without the involvement of an expert to inspect the damage or injury. Our goal is to enhance the simulated claims with a network structure similar to the one depicted in \Cref{fig:exampleSocialNetwork}(a). In this figure, the circles depict claims and the rectangles parties.

To accomplish this, we create a set $A_p$ for each party type, such as the set of garages $A_g$, brokers $A_b$, experts $A_e$, and other persons $A_o$. These sets represent the specific parties of each type within the larger set of all possible parties, denoted as $P = A_g \hspace{0.25mm} \cup \hspace{0.25mm} A_b \hspace{0.25mm} \cup \hspace{0.25mm} A_e \hspace{0.25mm} \cup \hspace{0.25mm} A_o$. The size of a specific set $A_p$ is determined by the corresponding user-specified parameter (see \Cref{subsec:PHConstractCharacteristics} and Appendix \ref{app:DefaultConfiguration}). For example, if the number of garages $n_{gar}$ is set to 150, the simulation engine will create a set $A_g$ of 150 unique garages. For each claim, we then randomly select one (or multiple when we connect the claim to other persons, see \Cref{tab:PHContractCharacteristics}) member from $A_p$ to link the claim to a specific party. By repeating this procedure for each type of party, we create a social network structure where every claim is connected to different (types of) parties (see \Cref{fig:exampleSocialNetwork}(a)). As a rule, we do not link the claim to an expert when $L_{ijk} < 250$ \citep{KBCBrussels}. 

Next, we proceed with generating the claim label. The data-generating fraud model is a logistic regression model
\begin{equation}\label{eq:ClaimLabel}
	\begin{aligned}
		Y_{ijk} \sim \text{Bern}\left( \pi_{ijk} \right) \hspace{2mm} \text{and} \hspace{2mm} \pi_{ijk} = \frac{e^{ {}_{0} \beta_f  + {}_{f} \bs{x}_{ijk}^\top \bs{\beta}_{f}}}{1 + e^{{}_{0} \beta_f + {}_{f} \bs{x}_{ijk}^\top \bs{\beta}_{f}}}. \\
	\end{aligned}
\end{equation}
Here, Bern denotes a Bernoulli distribution and $Y_{ijk}$ is a binary variable indicating if the $k^{th}$ claim of the $j^{th}$ contract of policyholder $i$ is fraudulent ($Y_{ijk} = 1$) or not ($Y_{ijk} = 0$). The subscript $f$ stands for fraud and ${}_{0} \beta_f$ is the intercept term. The relation between the fraud label and the features is adjusted via the covariate vector ${}_{f} \bs{x}_{ijk}$ and the parameter vector $\bs{\beta}_{f}$ (see Appendix \ref{app:ClaimLabel} for the default specification). 

We generate the claim labels in an iterative manner and this process is visualized in \Cref{fig:exampleSocialNetwork}. In this figure, the label of the claims is color-coded. Gray stands for unlabeled, red for fraudulent and green for non-fraudulent. Panel (a) represents the network at initialization, when all claims are unlabeled. We have no fraud-related information at this point and hence, no values for the social network features that rely on this information (e.g., the ratio of known fraudulent claims in the second order neighborhood). Consequently, at initialization, we remove all fraud-score and neighborhood based features (see \Cref{tab:SocialNetworkFeatures}) from ${}_{f} \bs{x}_{ijk}$ and $\bs{\beta}_{f}$ in \eqref{eq:ClaimLabel}. Next, we take a random subset, equal in size to 1\% for example, of all simulated claims. By deliberately taking a small subset of the data, we ensure that only a limited proportion of the claim labels is generated without effect of the fraud-score and neighborhood based features. In \Cref{fig:exampleSocialNetwork}(a), this subset consists of claims C1, C2 and C3. To generate the claim labels, we take a random draw from $\text{Bern}( \pi_{ijk})$ (see \eqref{eq:ClaimLabel}). As such, we generate our first set of labeled claims (see \Cref{fig:exampleSocialNetwork}(b)). This enables us to compute the values for the fraud-score and neighborhood based features. Consequently, from here on out, we include these features in ${}_{f} \bs{x}_{ijk}$ and $\bs{\beta}_{f}$. To generate the labels of the remaining claims, we again take a random subset of unlabeled claims. In \Cref{fig:exampleSocialNetwork}(b), this subset corresponds to claims C5 and C6. The size of this random subset can be set by the user. By default, this is equal to 10\% of all simulated claims. We combine this subset with the unlabeled claims in the $2^{nd}$ order neighborhood of the fraudulent claims in the previous iteration (i.e. C4 in \Cref{fig:exampleSocialNetwork}(b)). In doing so, we ensure that every subset includes unlabeled claims that are connected to fraudulent claims and that we propagate fraud through the network. We engineer the social network features for all claims in the subset and we take random draws from Bernoulli$(\pi_{ijk})$ to generate the claim labels (\Cref{fig:exampleSocialNetwork}(c)).

Algorithm \ref{algo:FraudGen} is a generalization of the process illustrated in \Cref{fig:exampleSocialNetwork}. We use this iterative algorithm to simulate the claim labels $Y_{ijk}$ and each iteration consists of three steps. First, we take a random subset of the unlabeled claims and we combine this subset with all unlabeled claims in the $2^{nd}$ order neighborhood of the fraudulent claims in the previous iteration. Second, we engineer the social network features for all claims in the subset. Third, we take random draws from Bernoulli$(\pi_{ijk})$ to generate the claim labels (\Cref{fig:exampleSocialNetwork}(c)). This concludes one iteration and we repeat the algorithm until all claims are labeled. 

\IncMargin{1em}
\LinesNotNumbered
\begin{algorithm}[!htbp]
	\SetAlgoLined
	\kwModel{$Y_{ijk} \sim \text{Bern}(\pi_{ijk})$}
	\kwInit{Remove fraud-score and neighborhood based features from $({}_{f} \bs{x}_{ijk},  \bs{\beta}_{f})$ in the first iteration and generate the initial claim labels using \eqref{eq:ClaimLabel}}
	\Repeat{all claims are labeled}{
		\nextnr
		Take a subset of the simulated database: a random sample of unlabeled claims combined with the unlabeled claims in $2^{nd}$ order neighborhood of fraudulent claims in the previous iteration\;\label{Step1}
		\nextnr
		Construct the social network features for the claims in this subset\; \label{Step2}
		\nextnr
		Generate the claim label using the data-generating logistic regression model in \eqref{eq:ClaimLabel}\; \label{Step3}
	}
	\caption{\label{algo:FraudGen}Iterative algorithm to simulate the claim labels}
\end{algorithm}

The user can specify which features are included in ${}_{f} \bs{x}_{ijk}$ and determine their effect size in $\bs{\beta}_{f}$. Consequently, the user has the flexibility to activate or deactivate specific feature effects and to control their strength. By including social network features in ${}_{f} \bs{x}_{ijk}$ and via the specified effect size in $\bs{\beta}_f$, for example, we determine to which extent the network exhibits patterns of homophily. The greater the corresponding effect size in $\bs{\beta}_{f}$, the more densely connected fraudulent claims will be. Conversely, we can turn off the social network effect by omitting the social network features from ${}_{f} \bs{x}_{ijk}$. Further, we can set the desired level of class imbalance. Hereto, our simulation engine determines which value for ${}_{0} \beta_f$ results in the target class imbalance (see Appendix \ref{app:ClaimLabel} for detailed information).

\subsection{Replicating the expert-based fraud detection approach}\label{subsec:Investigation}
In a real life fraud data set, we typically have historical, labeled data that is the result of an investigation by a fraud expert (see \Cref{subsec:FraudDetection}). In our simulation engine, we replicate the two steps of this fraud detection approach to obtain these labels. First, we flag claims as suspicious based on a set of business rules which can be defined by the user. Hence, an alert will be triggered for claims that meet the criteria outlined in the business rules. By default, we flag claim $k$ under contract $j$ of policyholder $i$ as suspicious if it satisfies one of the following criteria: a) the claim is filed within one year of the most recent claim (i.e. $\texttt{ClaimDate}_{ijk} - \texttt{ClaimDate}_{ij(k - 1)} \leq 1$); b) the individual claim amount $L_{ijk} >$ 75\% of \texttt{ValueCar}$_{ijk}$ or c) the cumulative claim amount $\sum_{l = 1}^k L_{ijl} >$ 200\% of \texttt{ValueCar}$_{ijk}$. In reality, these claims are passed to an expert who performs an in-depth investigation. Following the investigation, the expert judgement determines whether the claim is legitimate or not. We simulate the expert judgement in the second step as follows. For a claim that is flagged by the business rules in step one, we first look at its ground truth label $Y_{ijk}$. If $Y_{ijk}$ = \texttt{non-fraudulent}, we take a random draw from $Y_{ijk}^{\texttt{expert}} \sim \text{Bern}(0.01)$. Hence, when the claim is legitimate, we have a 99\% probability that the expert will label the claim as non-fraudulent. Conversely, if $Y_{ijk} =$ \texttt{fraudulent}, we randomly draw from $Y_{ijk}^{\texttt{expert}} \sim \text{Bern}(0.99)$. Thus, for fraudulent claims, there is a 99\% probability that the expert will classify them as fraudulent as well. Further, claims that are not flagged by the business rules obtain the label \texttt{uninvestigated}. Hereby, we create all three labels that are typically available in an insurance fraud data set: \texttt{non-fraudulent}, \texttt{fraudulent} or \texttt{uninvestigated}. By following the procedure as outlined above, we acknowledge and reflect the inherent missing information and uncertainties that exist in real-life data. That is, the expert-based approach is not entirely infallible \citep{Baesens2015}. Claims that are judged to be non-fraudulent by the investigation may in reality be fraudulent and vice versa. In addition, we acknowledge and replicate the phenomenon of having a substantial proportion of uninvestigated claims that are unlabeled.

%
%
%

%% file: Illustration.tex
\section{Generating synthetic fraud network data: illustrations}\label{sec:Illustration}

In this section, we illustrate the capabilities of our simulation engine. We specifically highlight the impact of social network features on the resulting synthetic data sets. Hereto, we generate and analyze two different types of data sets. \replaced[id = B]{In one data set}{One where} we include a moderately strong social network effect and \replaced[id = B]{in another one}{one where} we exclude it. Additionally, we provide an illustrative example of the construction and evaluation of a fraud detection model using a synthetically generated data set. We explore to which extent the constructed model is able to identify fraudulent claims that are not investigated and labeled by the expert.

\subsection{The impact of social network features on the synthetic data}
We generate two different types of data sets. In the first type of data set we introduce a moderately strong social network effect in the claim label generation (see Section \ref{subsec:Fraud}). We denote this type of data set as $\mc{D}^{Network}$. \Cref{tab:FraudModelSpecs} depicts the specification of the effect sizes used in the simulation of $\mc{D}^{Network}$. We include various types of features to generate a realistic and representative synthetic data set. These features encompass the policyholder, claim-specific, and social network characteristics. In order to replicate the social dynamics of fraud, we assign a strong effect size for the social network features \texttt{n1.size}, \texttt{n2.size} and \texttt{n2.ratioFraud}. Hereby, we create a synthetic data set where the network structure exhibits patterns of homophily. Conversely, in the second type of data set $\mc{D}^{Non-network}$, we exclude all network-related features from the data-generating fraud model (see \Cref{tab:FraudModelSpecs}). As such, we create a data set where fraud is not influenced by social interactions or network dynamics. The claim label generation is solely driven by policyholder and claim-specific characteristics. The selected set of policyholder and claim-specific features is identical in  $\mc{D}^{Network}$ and $\mc{D}^{Non-network}$, as well as the effect sizes of these features (see \Cref{tab:FraudModelSpecs}).

\begin{table}[!htbp]
	\centering
	\small
	\caption{Specification of the data-generating fraud model in $\mc{D}^{Network}$ and $\mc{D}^{Non-network}$. We generate the claim label $Y_{ijk}$ by taking a random draw from $\text{Bern}\left( \pi_{ijk} \right)$ where $\pi_{ijk} = \exp( {}_{0} \beta_f  + {}_{f} \bs{x}_{ijk}^\top \bs{\beta}_{f}) (1 + \exp({}_{0} \beta_f + {}_{f} \bs{x}_{ijk}^\top \bs{\beta}_{f}))^{-1}$.}
	\label{tab:FraudModelSpecs}
	\begin{tabular}{lcc}
		\hline
		& \multicolumn{2}{c}{$\beta_{f}$}\Tstrut\Cstrut\\
		\cline{2-3}
		Feature & $\mc{D}^{Network}$ & $\mc{D}^{Non-network}$ \Tstrut\Cstrut\\
		\hline
		Policyholder: &&\\
		\hspace{2mm} \texttt{AgePH} &-2.00&-2.00\\
		\hspace{2mm} \texttt{NrContractsPH} &-1.50&-1.50\\
		Claim-specific: &&\\
		\hspace{2mm} \texttt{ClaimAmount} &0.20&0.20\\
		\hspace{2mm} \texttt{ClaimAge} &-0.35&-0.35\\
		Social network: &&\\
		\hspace{2mm} \texttt{n1.size} &2.00&0\\
		\hspace{2mm} \texttt{n2.size} &-2.00&0\\
		\hspace{2mm} \texttt{n2.ratioFraud} &3.00&0\\
		\hline
	\end{tabular}
\end{table}

For both types of data sets, the number of policyholders is set to 200 000 and the target class imbalance (i.e., the ratio of the number of fraudulent claims to the total number of claims) to 2\%.  All other settings remain at their default values (see Appendix \ref{app:DefaultConfiguration}). We generate 100 data sets of each type and explore the distribution of the claim labels across these simulated data sets. Hereto, we calculate the frequency and relative frequency of the $Y_{ijk}$ categories (i.e., \texttt{fraudulent} and \texttt{non-fraudulent}) and $Y_{ijk}^{\texttt{expert}}$ (i.e., \texttt{fraudulent}, \texttt{non-fraudulent}, and \texttt{uninvestigated}) in each data set. The average, minimum, and maximum values for both the frequency and relative frequency are computed and presented in Table \ref{tab:DescrClaimLabels}. In all synthetic data sets, the empirical class imbalance is nearly identical to the target class imbalance. The minimum class imbalance in $\mc{D}^{Network}$ is 1.97\% and the maximum 2.06\%. In $\mc{D}^{Non-network}$, the minimum is 1.98\% and the maximum is 2.04\%. The class imbalance is also reflected in the expert judgement. Only a small fraction of claims are subject to investigation (approximately 9\%), and among those investigated, only a minority are found to be fraudulent. Further, the empirical distributions of both $Y_{ijk}$ and $Y_{ijk}^{\texttt{expert}}$ are similar across all simulated data sets.

\begin{table}[!htbp]
	\footnotesize
	\caption{\label{tab:DescrClaimLabels}The average, minimum and maximum frequency and relative frequency (\%) of the ground truth and expert-based claim labels across the 100 synthetic data sets.}
	\begin{tabular}{@{\extracolsep{1pt}}llccc@{}}
		\hline
		&& \multicolumn{3}{c}{Frequency (\%)} \Tstrut\Cstrut\\
		\cline{3-5}
		&& Average & Minimum & Maximum \Tstrut\Cstrut\\
		\hline
		\multirow{7}{*}{\rotatebox{90}{$\mc{D}^{Network}$}}
		&$Y_{ijk}$:&&&\\
		&\hspace{2mm} - Fraudulent & 2 175.37 (2.01\%) & 2 121.00 (1.97\%) & 2 219.00 (2.06\%) \\ 
		&\hspace{2mm} - Non-fraudulent & 105 937.66 (97.99\%) & 105 103.00 (97.94\%) & 106 655.00 (98.03\%) \\ 
		&$Y_{ijk}^{\texttt{expert}}$:&&&\\
		&\hspace{2mm} - Fraudulent & 284.28 (0.26\%) & 260.00 (0.24\%) & 318.00 (0.30\%) \\ 
		&\hspace{2mm} - Non-fraudulent & 9 149.38 (8.46\%) & 8 952.00 (8.30\%) & 9 431.00 (8.67\%) \\ 
		&\hspace{2mm} - Uninvestigated & 98 679.37 (91.27\%) & 98 021.00 (91.08\%) & 99 428.00 (91.45\%) \\ 
		\hline
		\multirow{7}{*}{\rotatebox{90}{$\mc{D}^{Non-network}$}}
		&$Y_{ijk}$:&&&\\
		&\hspace{2mm} - Fraudulent & 2 175.84 (2.01\%) & 2 139.00 (1.98\%) & 2 209.00 (2.04\%) \\ 
		&\hspace{2mm} - Non-fraudulent & 105 921.40 (97.99\%) & 104 978.00 (97.96\%) & 106 789.00 (98.02\%) \\ 
		&$Y_{ijk}^{\texttt{expert}}$:&&&\\
		&\hspace{2mm} - Fraudulent & 293.64 (0.27\%) & 254.00 (0.24\%) & 335.00 (0.31\%) \\ 
		&\hspace{2mm} - Non-fraudulent & 9 137.37 (8.45\%) & 8 924.00 (8.29\%) & 9 454.00 (8.72\%) \\ 
		&\hspace{2mm} - Uninvestigated & 98 666.23 (91.28\%) & 97 826.00 (91.01\%) & 99 366.00 (91.45\%) \\ 
		\hline
	\end{tabular}
\end{table}

\begin{figure}[!htbp]
	\centering
	\caption{\label{fig:EmpDistrDnetwork}Illustration of the features' empirical distribution in a synthetically generated $\mc{D}^{Network}$.}
	\makebox[\textwidth][c]{\includegraphics[width = \textwidth]{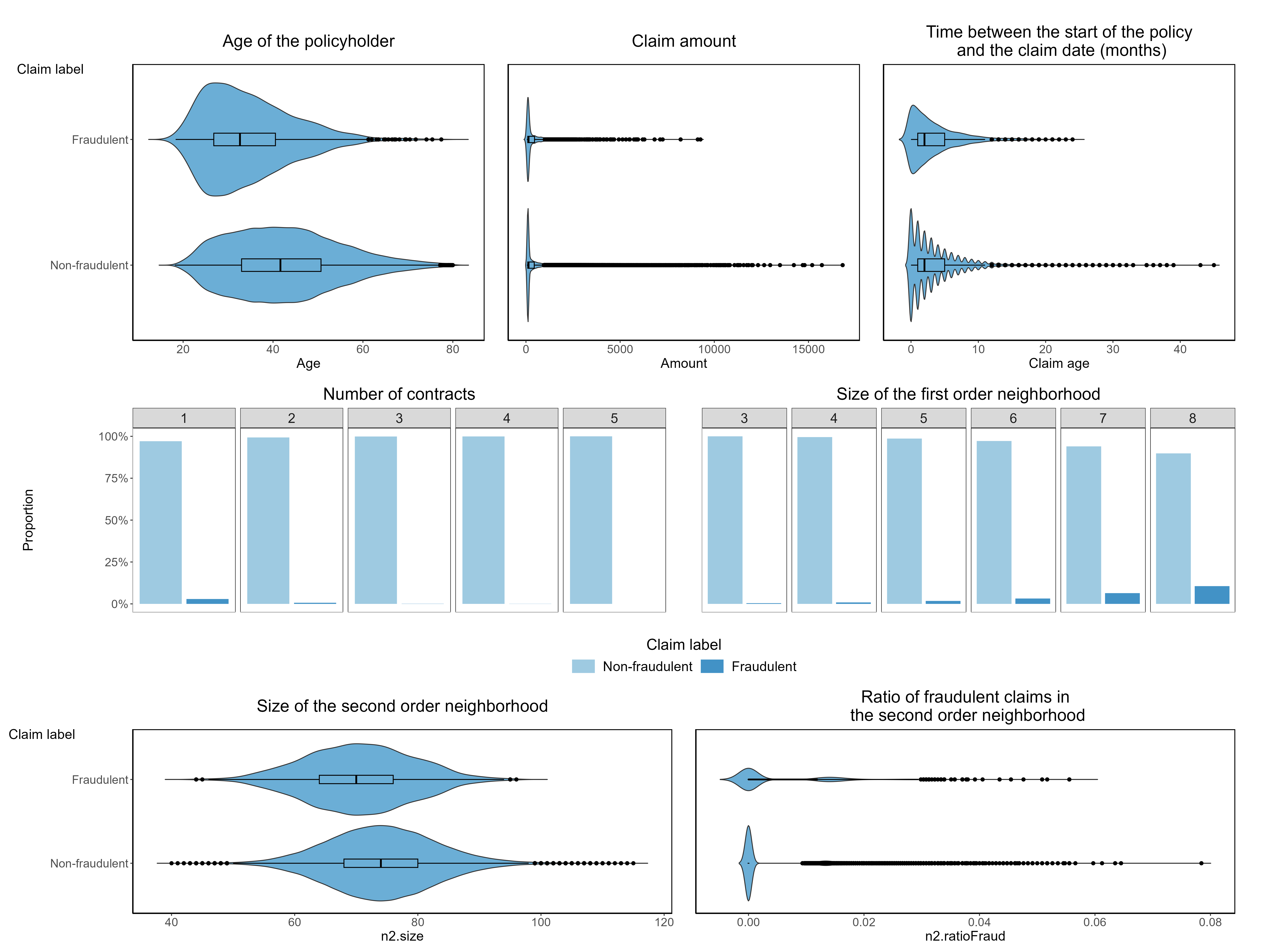}}
\end{figure}
\begin{figure}[!htbp]
	\centering
	\caption{\label{fig:EmpDistrDnonnetwork}Illustration of the features' empirical distribution in a synthetically generated $\mc{D}^{Non-network}$.}
	\makebox[\textwidth][c]{\includegraphics[width = \textwidth]{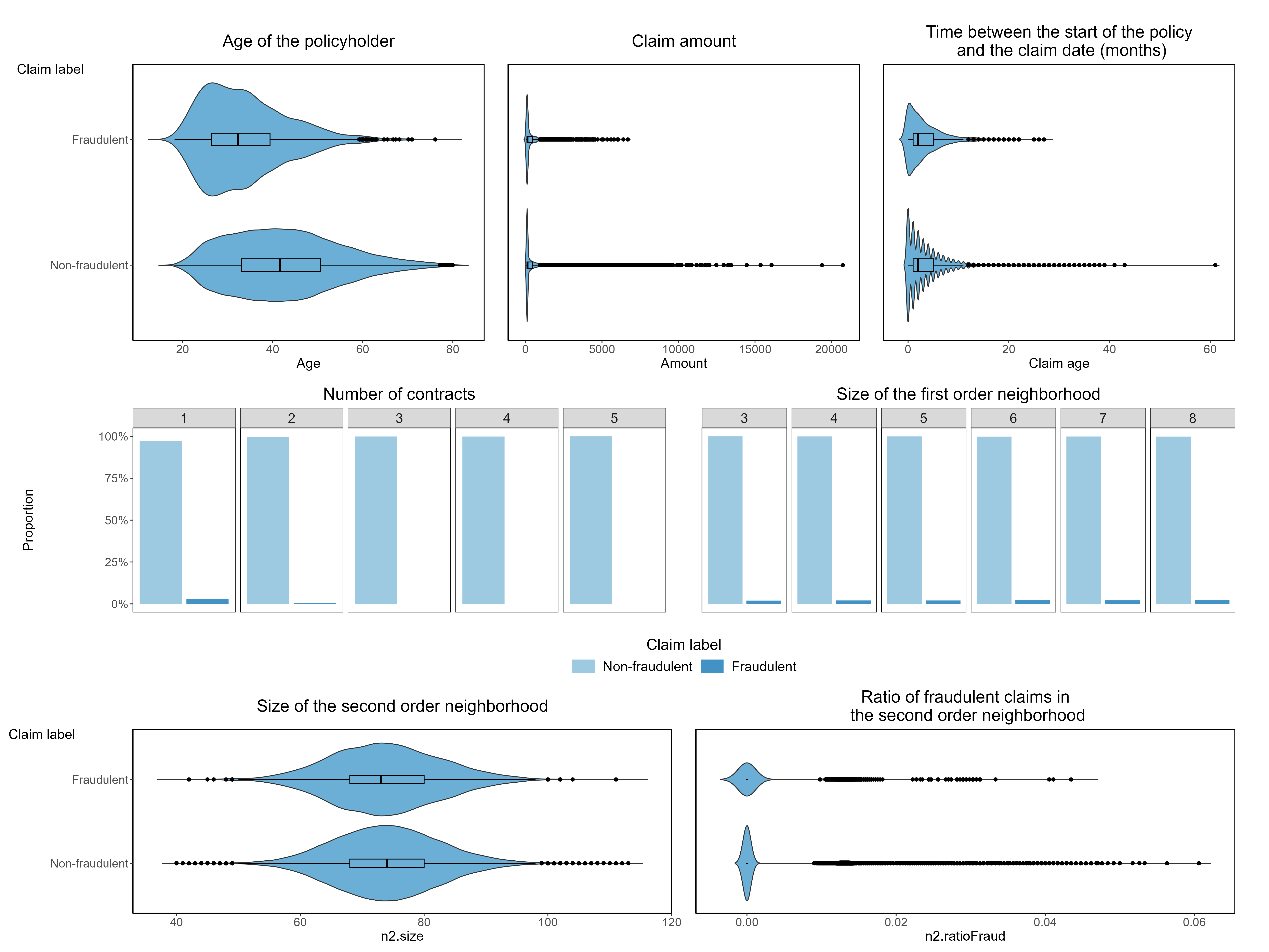}}
\end{figure}

\paragraph*{Empirical distribution of the features} \Cref{fig:EmpDistrDnetwork,fig:EmpDistrDnonnetwork} present the empirical distribution of the features included in the data-generating fraud models (see \Cref{tab:FraudModelSpecs}) of one synthetically generated data set. \Cref{fig:EmpDistrDnetwork} displays the features' empirical distribution in a simulated data set $\mc{D}^{Network}$. \Cref{fig:EmpDistrDnonnetwork} shows this in a synthetic data set $\mc{D}^{Non-network}$. The empirical distribution of policyholder and claim-specific features is different across fraudulent and non-fraudulent claims in both types of data sets. For example, fraudulent claims are mostly associated with younger policyholders (top left plot on \Cref{fig:EmpDistrDnetwork,fig:EmpDistrDnonnetwork}). Further, in $\mc{D}^{Network}$, the difference in the empirical distribution between fraudulent and non-fraudulent claims is also present for the social network features \texttt{n1.size}, \texttt{n2.size} and \texttt{n2.ratioFraud}. This suggests that the claim labels are linked to these features in $\mc{D}^{Network}$. In contrast, the empirical distributions of the social network features do not differ in $\mc{D}^{Non-network}$, indicating that there is no association between these features and the claim label.

\begin{figure}[!htbp]
	\centering
	\caption{\label{fig:Homophily}The empirical distribution of the dyadicity and heterophilicity in the synthetically generated $\mc{D}^{Network}$ and $\mc{D}^{Non-network}$ data sets.}
	\makebox[\textwidth][c]{\includegraphics[width = \textwidth]{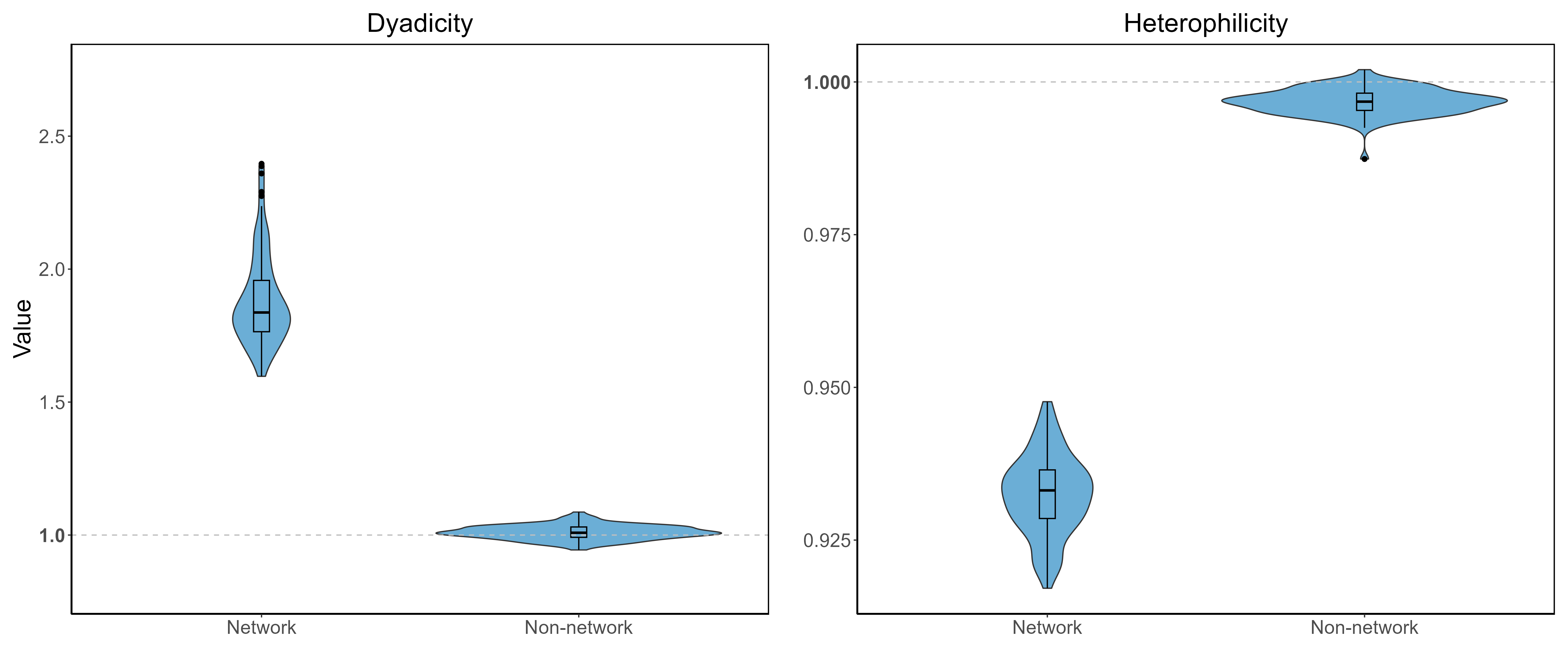}}
\end{figure}

\paragraph*{Homophily} \Cref{fig:Homophily} illustrates the dyadicity $\mathscr{D}$ and heterophilicity $\mathscr{H}$ observed in the simulated data sets (see \Cref{subsec:socialnetworkdata}). In $\mc{D}^{Network}$, the fraudulent claims are more densely connected to each other compared to what we expect by chance ($\mathscr{D} > 1$). In addition, fraudulent claims have fewer connections to non-fraudulent claims relative to what we expect by chance ($\mathscr{H} < 1$). In comparison, we observe no patterns of homophily in $\mc{D}^{Non-network}$. Both the dyadicity ($\mathscr{D} \approx 1$) and heterophilicity ($\mathscr{H} \approx 1$) correspond to values indicative of a random network configuration.

\begin{figure}[!htbp]
	\centering
	\caption{\label{fig:EstimatedCoefficients}Empirical distribution of the coefficient estimates across the (a) 100 simulated data sets $\mc{D}^{Network}$ and (b) 100 simulated data sets $\mc{D}^{Non-network}$. The red lines on the plot depict the features' effect size as specified in $\bs{\beta}_f$.}
	\makebox[\textwidth][c]{\includegraphics[width = \textwidth]{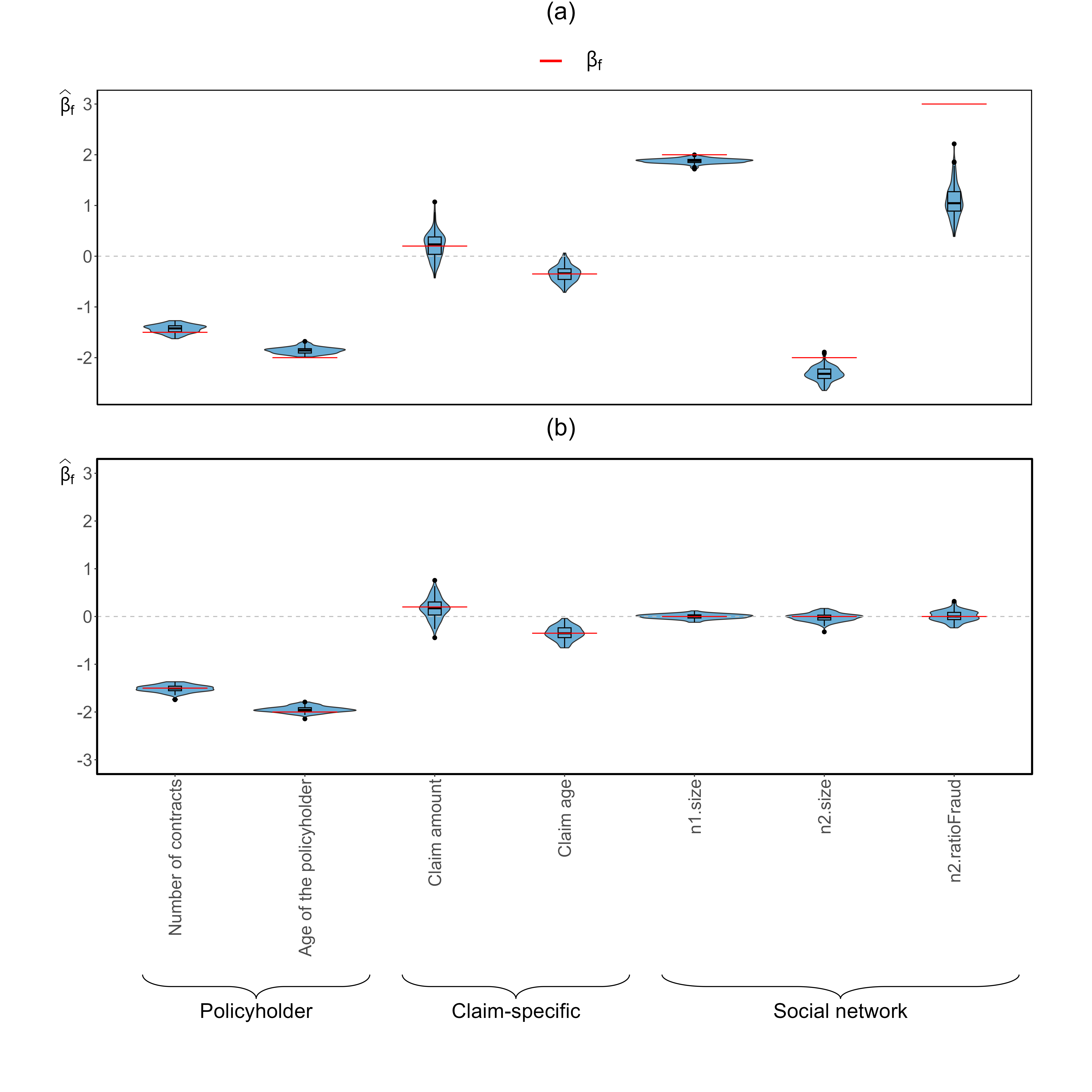}}
\end{figure}

\paragraph*{Effect size of the features} Next, we estimate the coefficient vector $\bs{\beta}_f$ in each synthetic data set. We fit the following logistic regression model
\begin{equation}\label{eq:DNetwork}
	\begin{aligned}
		\text{logit}(E[Y_{ijk}]) =& \ {}_{0} \beta_f + {}_{1} \beta_f \texttt{AgePH}_{i} + {}_{2} \beta_f \texttt{NrContractsPH}_{ij} +  {}_{3} \beta_f\texttt{ClaimAmount}_{ijk}\\
		&+  {}_{4} \beta_f\texttt{ClaimAge}_{ijk} + {}_{5} \beta_f \texttt{n1.size}_{ijk} + {}_{6} \beta_f \texttt{n2.size}_{ijk}\\
		&+ {}_{7} \beta_f \texttt{n2.ratioFraud}_{ijk}.\\
	\end{aligned}
\end{equation}
where $i$ refers to the policyholder, $j$ to the contract and $k$ to the claim. This model is the same as the data-generating fraud model (see \Cref{tab:FraudModelSpecs}). \Cref{fig:EstimatedCoefficients} depicts the empirical distribution of the estimated coefficient vector $\bs{\widehat{\beta}}$ across the 100 simulated data sets. Panel (a) shows the estimates obtained from $\mc{D}^{Network}$ and panel (b) from $\mc{D}^{Non-network}$. In both types of data sets, we observe some minor deviations from the specified effect size for most features, reflecting sampling variability. Further, the variability of the estimates is relatively small. Hence, we are able to accurately replicate the specified effect size for the different features, with only minor deviations due to sampling variability. The deviation from the specified effect size and variability, however, is substantially larger for the social network feature \texttt{n2.ratioFraud}. This feature represents the proportion of fraudulent claims in a claim's second order neighborhood. The estimated effect size of \texttt{n2.ratioFraud} is lower than its value as specified in $\bs{\beta}_f$. This is most likely attributable to the iterative growth in the number of fraudulent claims (see Algorithm \ref{algo:FraudGen}), leading to a deviation in the estimated effect size from the originally specified value. In the first iterations, there are only a few instances of fraudulent claims. Hence, most of the unlabeled claims will have similar values for \texttt{n2.ratioFraud}. As the number of iterations increases, there will be a progressive increase in the proportion of fraudulent claims (see Appendix \ref{app:n2.ratioFraud}). Accordingly, there will be more distinct feature values for \texttt{n2.ratioFraud}. Thus, the empirical distribution of \texttt{n2.ratioFraud} alters with each iteration.

\subsection{Exploring the capabilities of the simulation engine: evaluating a fraud detection model's effectiveness}
In this section, we illustrate the development and validation of a fraud detection model using a supervised learning technique (see \Cref{sec:FraudOverview}). We proceed with the 100 synthetic data sets $\mc{D}^{Network}$. In each simulated data set we construct a fraud detection model by fitting a logistic regression model to the investigated claims (i.e., those that are investigated and labeled by the expert, see \Cref{subsec:Investigation}). We rely on logistic regression given its robustness to imbalanced class sizes \citep{Oommen2011, Marques2013, Goorbergh2022}. \Cref{tab:DescrClaimLabels} illustrates that said imbalance is present in each synthetic data set. To examine the added value of social network analytics when a network effect is present, we define two distinct model specifications. For model 1, we only include policyholder and claim-specific features
\begin{equation}\label{eq:Model1}
	\begin{aligned}
		\text{logit}(E[Y_{ijk}]) =& \ {}_{0} \beta_f + {}_{1} \beta_f \texttt{AgePH}_{i} + {}_{2} \beta_f \texttt{NrContractsPH}_{ij} +  {}_{3} \beta_f\texttt{ClaimAmount}_{ijk}\\
		&+  {}_{4} \beta_f\texttt{ClaimAge}_{ijk}.\\
	\end{aligned}
\end{equation}
Next, we extend model 1 by incorporating social network features as well, resulting in model 2
\begin{equation}\label{eq:Model2}
	\begin{aligned}
		\text{logit}(E[Y_{ijk}]) =& \ {}_{0} \beta_f + {}_{1} \beta_f \texttt{AgePH}_{i} + {}_{2} \beta_f \texttt{NrContractsPH}_{ij} +  {}_{3} \beta_f\texttt{ClaimAmount}_{ijk}\\
		&+  {}_{4} \beta_f\texttt{ClaimAge}_{ijk} + {}_{5} \beta_f \texttt{n1.size}_{ijk} + {}_{6} \beta_f \texttt{n2.size}_{ijk}\\
		&+ {}_{7} \beta_f \texttt{n2.ratioFraud}_{ijk}.\\
	\end{aligned}
\end{equation}

To assess the predictive performance of the fraud detection models, we rely on the area under the receiver operating characteristic curve (AUC) \citep{AUC} and the top decile lift (TDL) \citep{Lemmens2006}. The AUC measures how well the model differentiates between fraudulent and non-fraudulent claims. An AUC of 0.5 corresponds to a random model and a perfect model has an AUC of 1. The TDL measures the extent to which a model surpasses a random model in detecting fraudulent claims. We calculate the TDL by dividing the proportion of fraudulent claims among the top 10\% of claims with the highest predicted probability by the relative frequency of fraudulent claims in the data set. The TDL of a random model is equal 1. The higher the TDL, the better the model performance. 

In each synthetic data set, we examine the in- and out-of-sample predictive performance of the fitted model. Hereto, we use the model fit to compute the probability of fraud for claims in the in- and out-of-sample data set
\begin{equation}
	\begin{aligned}
		\pi_{ijk} = \frac{e^{ {}_{0} \hat{\beta}_f  + {}_{f} \bs{x}_{ijk}^\top \bs{\widehat{\beta}}_{f}}}{1 + e^{{}_{0} \hat{\beta}_f  + {}_{f} \bs{x}_{ijk}^\top \bs{\widehat{\beta}}_{f}}}.
	\end{aligned}
\end{equation}
The in-sample data set consists of the investigated claims (approximately 9\% of all claims are investigated, see \Cref{tab:DescrClaimLabels}). Here, we use the labels of $Y_{ijk}^{\texttt{expert}}$ as outcome when computing the performance measures. The out-of-sample data set contains all uninvestigated claims. In our synthetic data set, we have the advantage of having access to the ground truth label $Y_{ijk}$ of the uninvestigated claims, which is not available in real-life data sets. We calculate the out-of-sample AUC and TDL using $Y_{ijk}$. As such, we evaluate to which extent our model is able to generalize and detect fraud in the unlabeled claims. 

\begin{figure}[!htbp]
	\centering
	\caption{\label{fig:PerfFDM}Distribution of the performance measures across the 100 simulated data sets $\mc{D}^{Network}$. The grey line in the plots corresponds to the performance of a random model. The in-sample performance is evaluated using the labels of the investigated claims. The out-of-sample performance is assessed using the ground truth label $Y_{ijk}$ of the uninvestigated claims.}
	\makebox[\textwidth][c]{\includegraphics[width = \textwidth]{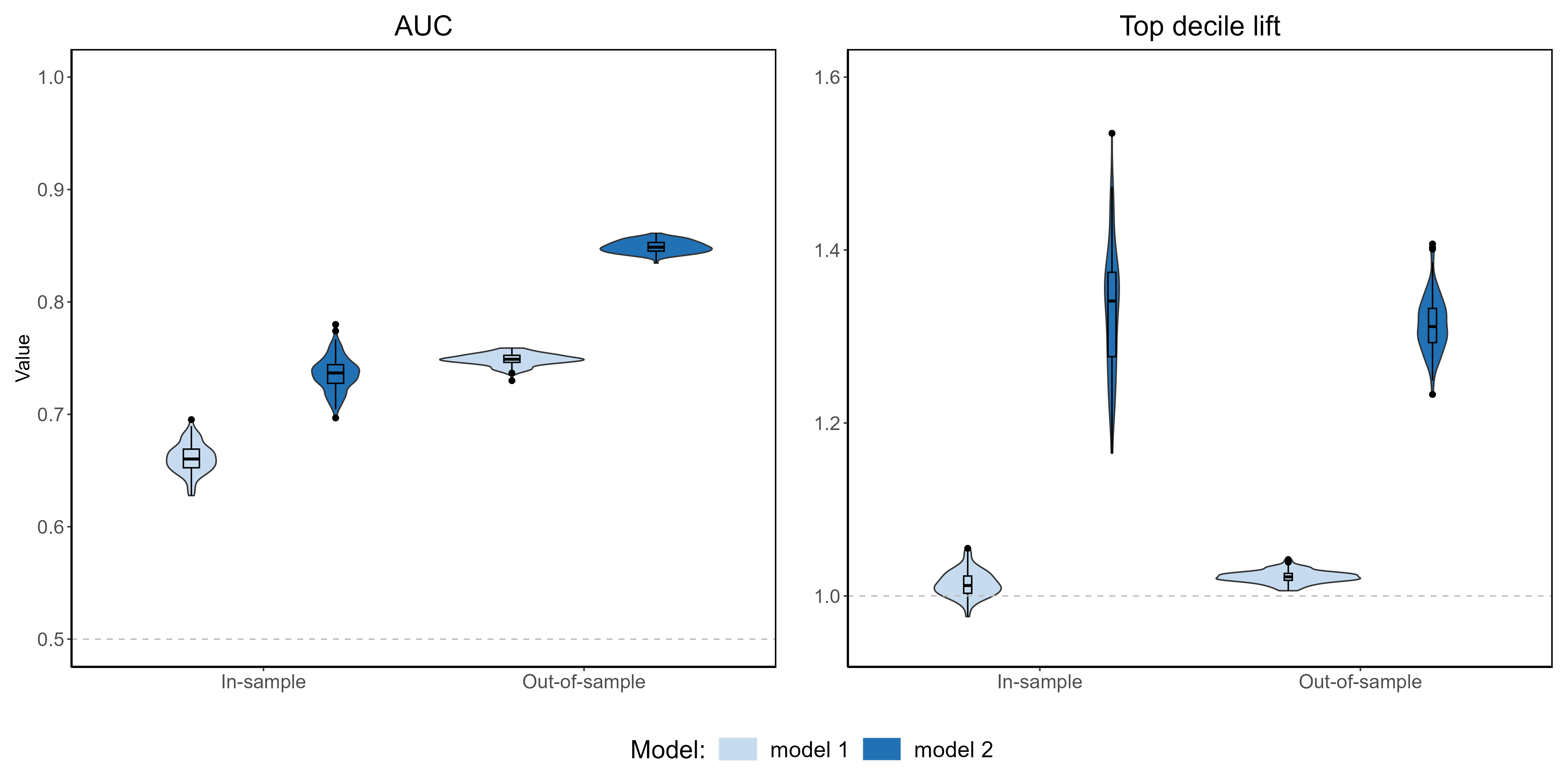}}
\end{figure}

\Cref{fig:PerfFDM} depicts the in- and out-of-sample predictive performance of model 1 and model 2. In terms of AUC and TDL, model 2 consistently outperforms model 1 in both the in- and out-of-sample evaluations. Consequently, by incorporating social network features in addition to the traditional claim characteristics, we enhance the model's ability to identify fraudulent claims. Furthermore, the TDL of model 1 approaches one in all simulated data sets, indicating that the model performs no better than random chance in identifying fraudulent claims within the top 10\% of predicted probabilities. In comparison, the TDL of model 2 is substantially larger than one. Model 2 also retains its predictive performance on the out-of-sample data sets. Hence, by training the model on the investigated claims, we can effectively capture the distinct patterns exhibited by fraudulent claims. One seemingly contradictory finding, however, is that the out-of-sample AUCs are higher than the in-sample AUCs. This discrepancy in performance is likely due to the different labels used for model evaluation. For the in-sample comparison, we rely on the expert judgment labels $Y_{ijk}^{\texttt{expert}}$. Conversely, for the out-of-sample comparison we use the ground-truth labels $Y_{ijk}$. This variation in label sources may contribute to the observed differences in model performance. In addition, the in-sample data sets are 10 times smaller than the the out-of-sample data sets. Consequently, the in-sample data sets exhibit more variability. Further, a small number of the investigated claims will be false positives or false negatives (see \Cref{subsec:Investigation}). When we fit the models with the ground truth-label $Y_{ijk}$ instead of $Y_{ijk}^{\texttt{expert}}$, the in-sample performance is higher compared to the out-of-sample performance (see Appendix \ref{app:PerformanceGroundTruth}).

%% file: Discussion.tex
\section{Discussion}\label{sec:Discussion}
In this paper, we present a powerful and flexible toolbox to generate synthetic insurance fraud network data. The simulation engine consists of seven consecutive steps which enable us to generate a complete and complex data set. The engine generates policyholder characteristics, contract-specific features, the number of claims and individual claim costs, and the claim labels (fraudulent or non-fraudulent). To ensure that the simulated data accurately reflects the real-world scenario, the fraud investigation process is also replicated. The generated data from each step is combined to produce a final synthetic database that can be used for various purposes. In generating the synthetic data, the user has control over various data-generating mechanisms. 

The simulation engine can produce diverse scenarios to meet different research needs. We showcase this ability by generating two distinct types of data sets, one where the social network effect is present during the claim label generation and one where it is absent. Our results highlight the toolbox's capability to simulate synthetic data according to the user-defined parameters. Our simulation engine accurately generates the desired class imbalance as well as the specified effect sizes of the  covariates (including the social network features). As such, we are able to generate data sets that closely mirror real-life insurance fraud data sets in motor insurance.

\added[id = B]{The synthetic datasets mirror the network structure and include covariates similar to those present in the real-life motor insurance fraud dataset analyzed in \citet{PaperMaria}. Our simulation engine, however, does not generate more granular information about the filed claim which might be available to the insurer. For instance, in real-life datasets, insurers often have access to additional information such as a written report or textual description accompanying the claim. Additionally, our simulation engine is tailored specifically for simulating synthetic motor insurance data. Future research could explore methods to generate finer detailed information on the claims and expanding the capabilities of the engine to cover other (non-)life insurance products.}

Researchers can utilize our simulation engine to conduct benchmark studies, aimed at addressing (methodological) challenges posed by insurance fraud. For instance, future research can focus on the evaluation of sampling techniques to handle the high class imbalance and the performance of learning methods in combination with said sampling techniques.

%% file: Acknowledgments.tex
\section*{Statements and Declarations}
No potential conflict of interest was reported by the authors.

Katrien Antonio gratefully acknowledges funding from the FWO and Fonds de la Recherche Scientifique - FNRS (F.R.S.-FNRS) under the Excellence of Science (EOS) programme, project ASTeRISK Research Foundation Flanders [grant number 40007517], from the Chaire DIALog by CNP Assurances and the FWO network W001021N. The authors gratefully acknowledge support from the Ageas research chair on insurance analytics at KU Leuven. Additionally, the authors would like to thank Misja Langeler for the assistance provided during the initial stages of the simulation engine.

%% file: Appendix.tex
\section{Default configuration of the simulation engine}\label{app:DefaultConfiguration}
The default settings for the database, policyholder, contract-specific and claim characteristics are given in \Cref{tab:DefaultConfiguration}. The default data-generating claim frequency and claim severity models are given in Appendix \ref{app:ClaimFreqSev}. In Appendix \ref{app:ClaimLabel}, we specify the default data-generating fraud model. 

We define the level of class imbalance as the ratio of the number of fraudulent claims to the total number of claims. Further, \texttt{ExcludeParties} is a parameter that allows to exclude certain types of parties from the network. We include all types of parties by default.
\clearpage
\begin{landscape}
\begin{table}[!htbp]
	\caption{Default configuration of the variables in the simulation engine.}
	\label{tab:DefaultConfiguration}
	\scriptsize
	\begin{threeparttable}
		\begin{tabular}{@{\extracolsep{0.75pt}}llHL{6.5cm}L{5.57cm}@{}}
			\toprule
			& Variable & Type & Description & Default value/generator\\
			\midrule
			\multirow{6}{*}{\rotatebox{90}{Data set}}
			&\texttt{TargetPrev} && Target level of class imbalance &\texttt{0.01}\\
			&\texttt{NrPH} && Number of policyholder & \texttt{10000}\\
			&\texttt{NrExperts} && Number of unique experts& \texttt{$\lfloor0.01\text{NrPH}\rfloor$}\\
			&\texttt{NrBrokers} &&Number of unique brokers& \texttt{$\lfloor0.01 \text{NrPH}\rfloor$}\\
			&\texttt{NrGarages} &&Number of unique garages& \texttt{$\lfloor0.03 \text{NrPH}\rfloor$}\\
			&\texttt{NrPersons} &&Number of unique persons& \texttt{$1.5 \text{NrPH}$}\\
			&\texttt{ExcludeParties} && Type of party to exclude & \texttt{Expert}\\
			\midrule
			\multirow{9}{*}{\rotatebox{90}{Policyholder}}
			&\texttt{AgePH} & Continuous & Age of the policyholder in years, default range is [18, 80] & $\mathcal{N}(40, 15)$ \\
			&\texttt{GenderPH} & Categorical & Gender of the policyholder, default settings are \texttt{female} if $u_i \leq 0.28$, \texttt{male} if $u_i > 0.29$ and \texttt{non-binary} otherwise& $u_i \sim U(0, 1)$\\
			&\texttt{ExpPH} & Continuous & Exposure of the policyholder in years, default range is [0, 20] & $\mathcal{N}(5, 1.5)$ \\
			&\texttt{RateNrContracts} & Continuous & Rate parameter $\lambda_i$ for generating \texttt{NrContractsPH} & $\lambda_i = 0.25 (1.05 - 2.5 \times 10^{-6} \times \text{\texttt{AgePH}}_i + 0.0025 \times \text{\texttt{AgePH}}_i^2 - 2.65 \times 10^{-5} \times \text{\texttt{AgePH}}_i^3)$ \\
			&\texttt{NrContractsPH} & Ordinal & Number of contracts. Default range is [1, 5]& $\text{Poi}(\lambda_i)$\\
			\midrule
			
			\multirow{10}{*}{\rotatebox{90}{Contract-specific}}
			&\texttt{ExpPHContracts} & Continuous & Exposure corresponding to the contract & \texttt{ExpPH}$_{i} - $ U$(0, \text{\texttt{ExpPH}}_{i} / 2)$ if \texttt{NrContractsPH}$_{i} > 1$,\\
			&&&& \hspace{0.5em}else \texttt{ExpPHContracts}$_{ij} = $ \texttt{ExpPH}$_{i}$\\
			&\texttt{AgeCar} & Continuous & Age of the vehicle in years & $\max(\mathcal{N}(7.5, \sqrt{5}), \text{\texttt{ExpPHContracts}}_{ij})$\\
			&\texttt{OrigValueCar} & Continuous & Original value of the vehicle & $\text{Exp}(\lambda_i / \text{NrContractsPH}_i)$ \\
			&\texttt{ValueCar} & Continuous & Current value of the car & $\text{\texttt{OrigValueCar}}_{ij} (1 - \delta\tnote{a} \ )^{\text{\texttt{AgeCar}}_{ij}}$\\
			&\texttt{Coverage} & Categorical & Type of coverage provided by the insurance company (see Appendix \ref{app:Coverage}) & $\text{Multinomial}(1, \pi_{\text{TPL}}, \pi_{\text{PO}}, \pi_{\text{FO}})$\\
			&\texttt{Fuel} & Categorical & Type of fuel of the vehicle & Bernoulli$(0.3)$\\
			&\texttt{BonusMalus} & Ordinal & Level occupied in bonus-malus scale of the insurance company & $\min(\lfloor \mc{G}(1, 1 / 3) \rfloor$, 22) \\
			\midrule
			
			\multirow{8}{*}{\rotatebox{90}{Claim}}
			& \texttt{ClaimAge} & Integer & Number of months from the contract's inception to the date of the incident & $\lfloor \text{Exp}(0.25) \rfloor$\\
			& \texttt{ClaimDate} & Continuous & Number of years between the start of the contract & $\max(U(0, \text{\texttt{ExpPHContracts}}_{ij}),$\\
			&&& and the claim's filing date & \hspace{1.75em} $\texttt{ClaimAge}_{ijk} / 12)$\\
			&\texttt{Police} & Categorical & Whether police was called when the incident happened & Bernoulli$(0.25)$\\
			&\texttt{nPersons} & Integer & Number of people involved in the claim, range is [0, 5] &  $S \xleftarrow{\pi_p} x$, $S = (0, 1, 2, 3, 4, 5)$ and $\pi_p = \textcolor{black}{(0.02, 0.57, 0.19, 0.10, 0.10, 0.02)}$\\
			
			\bottomrule
		\end{tabular}
		\begin{tablenotes}
			\item[a] If \texttt{OrigValueCar} $< 30 000$, $\delta = 0.15$ and $\delta = 0.075$ otherwise. Hence, more expensive cars have a lower depreciation rate.
		\end{tablenotes}
	\end{threeparttable}
\end{table}
\end{landscape}


\section{Limiting the range of feature values}\label{app:LimitRange}
\added[id=B]{For \texttt{AgePH}, the user can define a range of acceptable values $[\min_{\texttt{AgePH}}, \max_{\texttt{AgePH}}]$ where $\min_{\texttt{AgePH}}$ denotes the lower bound and $\max_{\texttt{AgePH}}$ the upper bound. The default range is $[18, 80]$. Using the generator, we simulate $\texttt{AgePH}_i$ for $i = (1, \dots, n_{ph})$ (see \Cref{tab:PHContractCharacteristics,tab:DefaultConfiguration}). Hereafter, we calculate the relative frequency for all values $\lceil \texttt{AgePH}_i \rceil$ that fall within the predefined range $[\min_{\texttt{AgePH}}, \max_{\texttt{AgePH}}]$. The ceiling function $\lceil \cdot \rceil$ is used to round values to the nearest integer. Values outside of the range $[\min_{\texttt{AgePH}}, \max_{\texttt{AgePH}}]$ are redistributed among the integer values falling within this interval, proportional to the relative frequency of the values within this range. Hereafter, we take a random draw from $U(0, 1)$ and add this value to the integer to obtain a numeric value.}

\textcolor{black}{The default range $[\min_{\texttt{ExpPH}}, \max_{\texttt{ExpPH}}]$ for \texttt{ExpPH} is set to $[0, 20]$.} Furthermore, \texttt{AgePH}\textcolor{black}{$_{i}$} - \texttt{ExpPH}\textcolor{black}{$_{i}$} cannot be smaller than the user-specified \textcolor{black}{$\min_{\texttt{AgePH}}$}. This would imply that the contract started before the policyholder \replaced[id = B]{reached the}{is} legal age of driving. Hereto, we define \texttt{MaxExp}$_i = $ \texttt{AgePH}$_i$ - \textcolor{black}{$\min_{\texttt{AgePH}}$}. For values outside the prespecified range, we redraw a value from $U$(\textcolor{black}{$\min_{\texttt{ExpPH}}$}, \texttt{MaxExp}$_i$).

We define $[1, 5]$ as default range for \texttt{NrContractsPH}. Values outside this range are rounded to the closest boundary.

\section{Simulating type of coverage}\label{app:Coverage}
The type of coverage is a nominal variable with three levels. We rely on a multinomial regression model to generate the type of coverage as a function of \texttt{ValueCar}, \texttt{AgeCar} and \texttt{AgePH}. The general form of the multinomial logistic regression model \citep{Agresti2013} is

$$ \log \left( \frac{\pi_{j}(\bs{x}_i)}{\pi_{J}(\bs{x}_i)} \right) = \bs{x}_i^\top \bs{\beta}_j, \hspace{1em} j = (1, \dots, J - 1), $$

where $\pi_{j}(\bs{x}_i) = P(\texttt{Coverage}_i = j | \bs{x}_i)$ denotes the probability that $\texttt{Coverage}_i$ equals category $j$. $\texttt{Coverage}_i$ denotes the response variable's value for observation $i$ and here, we use $j = (1, \dots, J)$ as an index for the categories of the response variable. We use category $J$ as reference category. $\bs{x}_{i}$ denotes the covariate vector and $\bs{\beta}_j$ is the parameter vector for category $j$. For notational simplicity, we assume that $\bs{x}_i$ is fixed for all categories $j = (1, \dots, J - 1)$. Further, $\sum_{j = 1}^J \pi_j(\bs{x}_i) = 1 \hspace{0.5em} \forall \hspace{0.5em} i \in (1, \dots, N)$ where \pdfmarkupcomment[markup=Highlight,color=yellow]{$N$}{Notatie niet eerder ingevoerd. Enkel voor number of policyholders, number of contracts per policyholder en number of claims per contract.} denotes the total number of observations.

The category-specific probability for category $j = (1, \dots, J - 1)$ is calculated as
$$ \pi_{j}(\bs{x}_i) = \frac{e^{\bs{x}_i^\top \bs{\beta}_j}}{1 + \sum_{h = 1}^{J - 1} e^{\bs{x}_i^\top \bs{\beta}_h}}$$
and we calculate this probability for reference category $J$ as
$$\pi_{J}(\bs{x}_i) = \frac{1}{1 + \sum_{h = 1}^{J - 1} e^{\bs{x}_i^\top \bs{\beta}_h}}.$$

\noindent
In our simulation engine, $\texttt{Coverage}_{i} \in$ (\texttt{TPL}, \texttt{PO}, \texttt{FO}) and we define
\begin{equation}\label{eq:multinomial}
	\begin{aligned}
		\bs{\beta}_{\text{\texttt{TPL}}} &= (\log(0.50), \log(1.25), \log(0.25)),\\
		\bs{\beta}_{\text{\texttt{PO}}} &= (\log(1.25), \log(0.75), \log(1.05)),\\
		\bs{\beta}_{\text{\texttt{FO}}} &= (\log(1.50), \log(0.75), \log(1.25)).\\
	\end{aligned}
\end{equation}
\noindent
The covariate vector $\bs{x}_i$ consists of the normalized values for the value of the car, age of the car and age of the policyholder. Given a variable $a = (a_1, \dots, a_i, \dots, a_N)$, we normalize \replaced[id = B]{it to the range}{in} $[-1, 1]$ using
\begin{equation}\label{eq:Standardizing}
	\begin{aligned}
		2 \frac{(a - \min(a))}{\max(a) - \min(a)} - 1.
	\end{aligned}
\end{equation}
\noindent

Hereafter, we calculate the probabilities $(\pi_{\text{TPL}}(\bs{x}_i), \pi_{\text{PO}}(\bs{x}_i), \pi_{\text{FO}}(\bs{x}_i))$ for all observations. We generate values for the type of coverage by taking random draws from $\text{Multinomial}(1, \pi_{\text{TPL}}, \pi_{\text{PO}}, \pi_{\text{FO}})$.

Using the default values (see equation \eqref{eq:multinomial}), the probability of signing up for a full omnium is larger for expensive, relatively new cars and older policyholders. Similarly, the probability of taking out a partial omnium is higher for expensive, relatively new cars but here the effect of age is less strong. Young policyholders with an inexpensive, older car have a higher probability to take out a policy with only third party liability.

\section{Claim frequency and claim severity model}\label{app:ClaimFreqSev}
Both the claim frequency and claim severity model are based on the results in \citet{Henckaerts2018}. In this paper, the authors fit a claim frequency and claim severity model on a motor insurance portfolio from a Belgian insurer. Further, \citep{Henckaerts2018} used a data-driven method to bin the continuous variables \texttt{AgePH}, \texttt{AgeCar} and \texttt{BonusMalus} into categorical variables. These bins are given in \Cref{tab:bins}. We denote these binned versions as \texttt{AgePHBin}, \texttt{AgeCarBin} and \texttt{BonusMalusBin}. By default, we use these binned versions in the data-generating claim frequency and claim severity model (see \Cref{tab:DefaultClaimModels}). 

\begin{table}[H]
	\centering
	\small
	\caption{\label{tab:bins}Bins of the continuous variables as used in \citet{Henckaerts2018}.}
	\begin{tabular}{ll}
		\hline
		Variable & Bins \\
		\hline
		\texttt{AgePHBin} & $[18,26]; (26,30]; (30,36]; (36,50]; (50,60]; (60,65]; (65,70];$\\
		& $(70,80]$\\
		\texttt{AgeCarBin} & $[0,5]; (5,10]; (10,20]; (20,\text{max(\texttt{AgeCar})}]$\\
		\texttt{BonusMalusBin} & $[0,1); [1,2); [2,3); [3,7); [7,9); [9,11); [11,22]$\\
		\hline
	\end{tabular}
\end{table}
\begin{table}[H]
	\centering
	\footnotesize
	\caption{Default specification of the claim frequency and claim severity model.}
	\label{tab:DefaultClaimModels}
	\begin{tabular}{lP{2cm}P{2cm}}
		\hline
		Variable & $\beta_{cf}$ & $\beta_{cs}$ \\ 
		\hline
		(Intercept) & -2.18 & 6.06 \\ 
		\texttt{AgePH}: &&\\
		\hspace{2mm}[18,26] (reference) &  & \\ 
		\hspace{2mm}(26,30] & $\log(0.85)$ & $\log(0.85)$\\ 
		\hspace{2mm}(30,36] & $\log(0.75)$ & $\log(0.75)$\\ 
		\hspace{2mm}(36,50] & $\log(0.70)$  & $\log(0.85)$\\ 
		\hspace{2mm}(50,60] & $\log(0.60)$ & $\log(0.85)$\\ 
		\hspace{2mm}(60,65] & $\log(0.55)$  & $\log(1.15)$\\ 
		\hspace{2mm}(65,70] & $\log(0.60)$ & $\log(1.25)$\\
		\hspace{2mm}(70, max(\texttt{AgePH})] & $\log(0.70)$ & $\log(1.50)$ \\ 
		
		
		\texttt{Coverage}:&&\\
		\hspace{2mm}\texttt{TPL} (reference) &&\\
		\hspace{2mm}\texttt{PO} & -0.12 & -0.16\\ 
		\hspace{2mm}\texttt{FO} & -0.11 & 0.11\\ 
		
		\texttt{AgeCarBin}:&&\\
		\hspace{2mm}(0, 5] (reference)&&\\
		\hspace{2mm}(5,10]&$\log(0.90)$& 0\\
		\hspace{2mm}(10,20]&$\log(0.80)$& 0\\
		\hspace{2mm}$(20,\text{max(\texttt{AgeCar})}]$&$\log(0.60)$& 0\\
		
		\texttt{Fuel}: &&\\
		\hspace{2mm}\texttt{Gasoline/LPG/Other} (reference) &&\\
		\hspace{2mm}\texttt{Diesel} & $\log(1.19)$ & 0\\ 
		
		\texttt{BonusMalusBin}: &&\\
		\hspace{2mm}[0,1) (reference) &&\\
		\hspace{2mm}[1,2) & 0.12 & 0.10\\ 
		\hspace{2mm}[2,3) & 0.18 & 0.15\\ 
		\hspace{2mm}[3,7) & 0.34 & 0.15\\ 
		\hspace{2mm}[7,9) & 0.48 & 0.15\\ 
		\hspace{2mm}[9,11) & 0.54 & 0.20\\ 
		\hspace{2mm}[11,22] & 0.78 & 0.30\\ 
		\hline
	\end{tabular}
\end{table}

\section{Data generating fraud model and class imbalance}\label{app:ClaimLabel}
\Cref{tab:DefaultFraud} depicts the default specification of the data-generating fraud model. The column names indicate which features are included, while the values represent the corresponding value in $\bs{\beta}_f$. Further, in the data-generating model we use the normalized version of the features \texttt{ClaimAmount}, \texttt{ClaimAge}, \texttt{n1.size}, \texttt{n2.size}, \texttt{AgePH} and \texttt{n2.ratioFraud} (see \eqref{eq:Standardizing}). Hereby, we bring all features to the same scale. This ensures that the features' effect sizes, as specified in $\bs{\beta}_f$, are comparable. Furthermore, at the end of every iteration in Algorithm \ref{algo:FraudGen}, we normalize both \texttt{n2.ratioFraud} and \texttt{n2.ratioNonFraud}. We do so since the network grows with every step in the algorithm. By normalizing these features, we aim to mitigate the influence of fluctuating values across iterations (also see Appendix \ref{app:n2.ratioFraud}).

\begin{table}[!htbp]
	\centering
	\scriptsize
	\caption{\label{tab:DefaultFraud}Default specification of the data-generating fraud model.}
	\begin{tabular}{lccccccc}
		\hline
		& \texttt{ClaimAmount} & \texttt{ClaimAge} & \texttt{n1.size} & \texttt{n2.size} & \texttt{NrContractsPH} & \texttt{AgePH} & \texttt{n2.ratioFraud}\\
		\hline
		$\bs{\beta}_{f}$ & 0.20 & -0.35 & 2.00 & -2.00 & -1.50 & -2.00 & 3.00\\
		\hline
	\end{tabular}
\end{table}

Further, the simulation engine allows us to specify the desired level of class imbalance $p_t$. We achieve this by employing the following approach in the third step of Algorithm \ref{algo:FraudGen}, where we generate the claim labels 
\begin{equation}
	\begin{aligned}
		Y_{ijk} \sim \text{Bern}\left( \pi_{ijk} \right) \hspace{2mm} \text{and} \hspace{2mm} \pi_{ijk} = \frac{e^{ {}_{0} \beta_f  + {}_{f} \bs{x}_{ijk}^\top \bs{\beta}_{f}}}{1 + e^{{}_{0} \beta_f + {}_{f} \bs{x}_{ijk}^\top \bs{\beta}_{f}}}. \\
	\end{aligned}
\end{equation}
Before generating $Y_{ijk}$, we set a seed for the random number generator to ensure reproducibility. We achieve the desired level of imbalance $p_t$ by optimizing
\begin{equation}\label{eq:Imbalance}
	\begin{aligned}
		\min_{{}_{0} \beta_{f}} |p_t - {}_{a} p|
	\end{aligned}
\end{equation}
\noindent
\added[id = B]{where the range of possible values for $_{0} \beta_f$ is set to [-10, 10]} and where
\begin{equation}
	\begin{aligned}
		{}_{a} p = \frac{\sum_{i, j, k} I(Y_{ijk} = \texttt{fraudulent})}{\sum_{i, j, k} I(Y_{ijk} = \texttt{fraudulent} \ \text{OR} \ Y_{ijk} = \texttt{non-fraudulent})}
	\end{aligned}
\end{equation}
\noindent
denotes the actual level of class imbalance in the synthetic data set (using all available claim labels). Here, $I(\cdot)$ represents the indicator function. \added[id = B]{However, for certain randomly drawn sets of $Y_{ijk}$, it is possible that there is no optimal ${}_{0} \beta_f$ (i.e., when ${}_{0} \beta_f$ is equal to -10 or 10) in a given iteration. In this scenario, we change the seed and generate another set of $Y_{ijk}$. We return to step 1 of Algorithm \ref{algo:FraudGen} when no optimal ${}_{0} \beta_f$ is identified following 100 seed changes.}

\section{Distribution values n2.ratioFraud}\label{app:n2.ratioFraud}
\begin{figure}[H]
	\centering
	\caption{\label{fig:n2.ratioFraud}Distribution of the values of \texttt{n2.ratioFraud} in each iteration. Per iteration, the violin plot depicts the density of \texttt{n2.ratioFraud} in the subset.}
	\makebox[\textwidth][c]{\includegraphics[width = \textwidth]{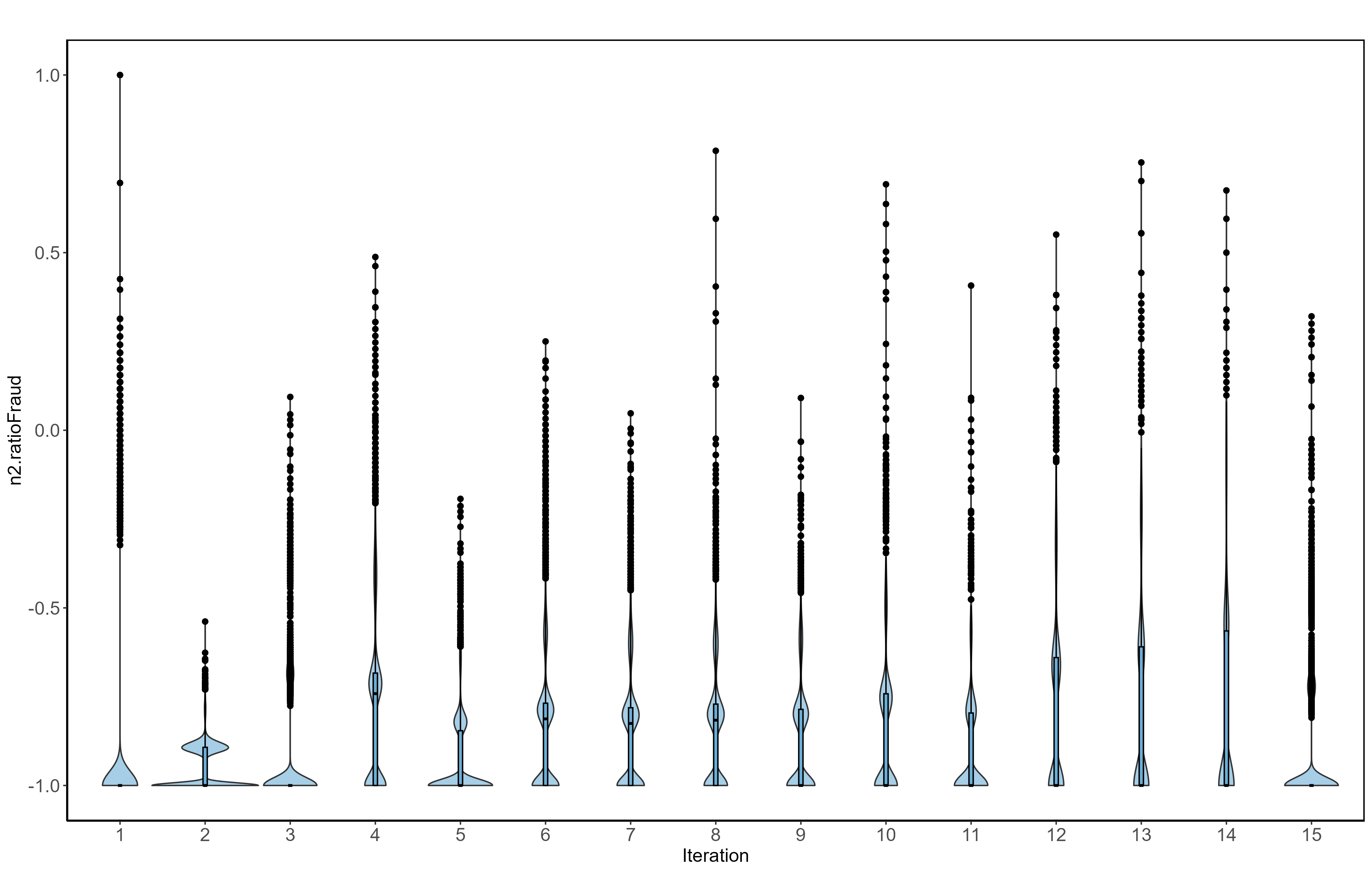}}
\end{figure}

\Cref{fig:n2.ratioFraud} depicts the features values of \texttt{n2.ratioFraud} across the different iterations of Algorithm \ref{algo:FraudGen} when generating the claim labels in \pdfmarkupcomment[markup=Highlight,color=yellow]{a}{is in 1 specifieke data set, niet over 100 data sets heen} synthetic data set. The horizontal axis depicts the iteration and the vertical axis the feature values. Per iteration, we show the empirical distribution of \texttt{n2.ratioFraud} in the random subset (see Algorithm \ref{algo:FraudGen}). In the first iterations, we have a small number of fraudulent claims. As a consequence, most observations have similar feature values for \texttt{n2.ratioFraud}. The scarcity of distinct values is evident from the compactness of the violin plots. With each iteration, the number of fraudulent claims grows, resulting in an increase in distinct values for \texttt{n2.ratioFraud}. In \Cref{fig:n2.ratioFraud}, this is reflected by the increase in width of the violin plots.

\section{Predictive performance in the synthetic data sets}\label{app:PerformanceGroundTruth}
\begin{figure}[!htbp]
	\centering
	\caption{\label{fig:PerfFDM_GroundTruth}Distribution of the performance measures in $\mc{D}^{Network}$. The grey line in the plots corresponds to the performance of a random model. The in-sample performance is evaluated using the ground truth label $Y_{ijk}$ of the investigated claims. The out-of-sample performance is assessed using the ground truth label $Y_{ijk}$ of the uninvestigated claims.}
	\makebox[\textwidth][c]{\includegraphics[width = \textwidth]{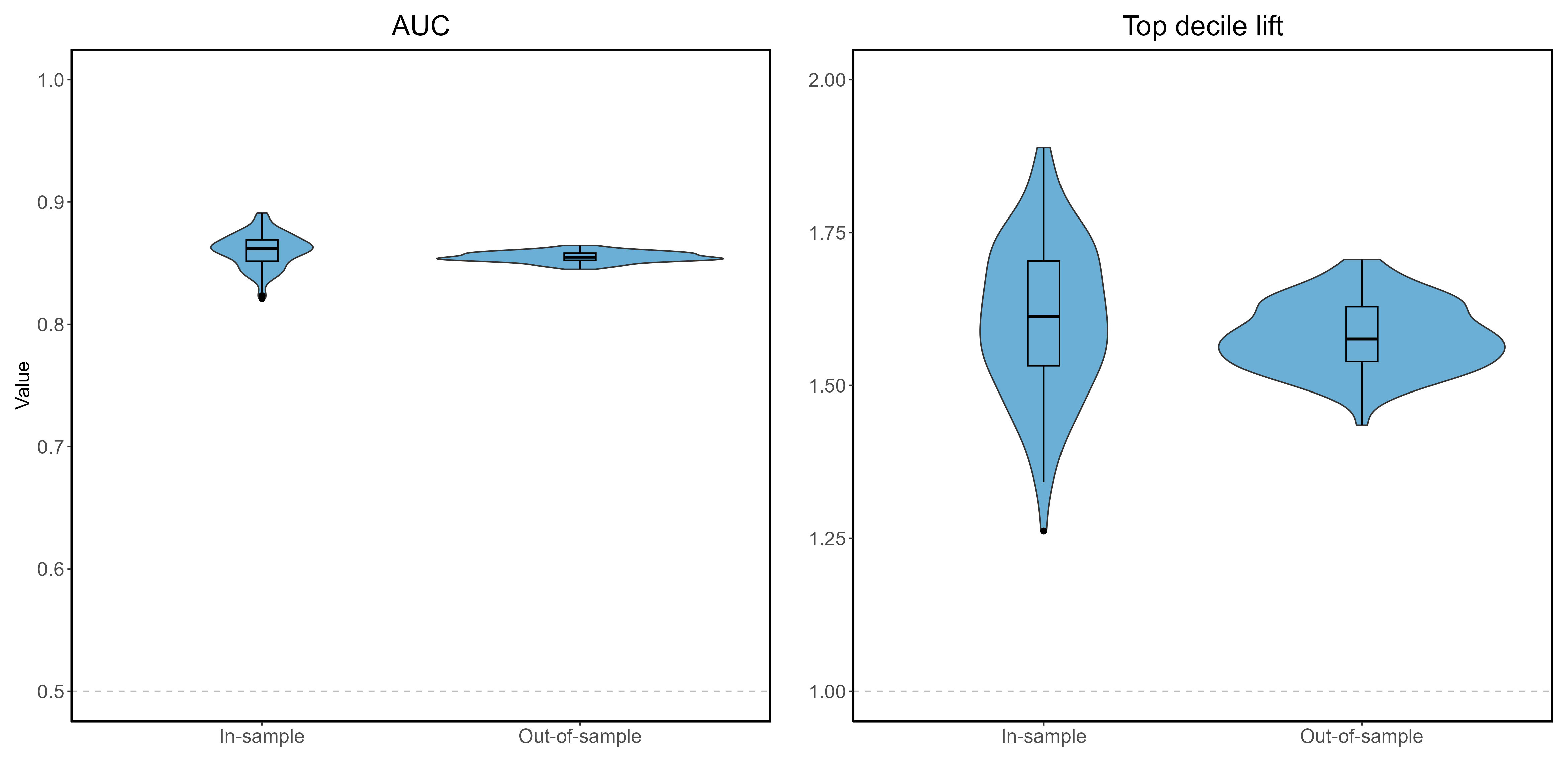}}
\end{figure}

To obtain insights into the results in \Cref{sec:Illustration}, we first fit the following \pdfmarkupcomment[markup=Highlight,color=yellow]{model}{Ik gebruik hier 1 modelspecificatie ter simplificatie. Main aim is om inzicht te verwerven in het verschil tussen de performantie maten op de in- en out-of-sample set, wat deels beïnvloed wordt door de response die we gebruiken.} to the investigated claims
\begin{equation}
	\begin{aligned}
		\text{logit}(E[Y_{ijk}]) &= \ {}_{0} \beta_f + {}_{1} \beta_f \texttt{AgePH}_{i} + {}_{2} \beta_f \texttt{NrContractsPH}_{ij} +  {}_{3} \beta_f\texttt{ClaimAmount}_{ijk}\\
		&+  {}_{4} \beta_f\texttt{ClaimAge}_{ijk} + {}_{5} \beta_f \texttt{n1.size}_{ijk} + {}_{6} \beta_f \texttt{n2.size}_{ijk}\\
		&+ {}_{7} \beta_f \texttt{n2.ratioFraud}_{ijk}.\\
	\end{aligned}
\end{equation}
\noindent
Hence, here we use the ground-truth $Y_{ijk}$ instead of the expert judgement $Y_{ijk}^{\texttt{expert}}$ as the response variable. Hereafter, we examine the performance on the investigated (i.e., the in-sample data set) and uninvestigated (i.e., the out-of-sample data set) claims. \Cref{fig:PerfFDM_GroundTruth} depicts the performance on the in- and out-of-sample data sets. Both the AUC and TDL indicate that the fitted models are substantially better than a random model. In addition, compared to the in-sample data sets, the performance is slightly lower on the out-of-sample data sets. Hence, the model is able to capture the underlying relationships between the predictors and the target variable.

